\documentclass[conference]{IEEEtran}
\IEEEoverridecommandlockouts

\usepackage{cite}
\usepackage{amsmath,amssymb,amsfonts}
\usepackage{graphicx}
\usepackage{booktabs}
\usepackage{array}
\usepackage{multirow}
\usepackage{algorithm}
\usepackage{algpseudocode}
\usepackage{xcolor}
\usepackage{url}
\usepackage{tikz}
\usetikzlibrary{positioning,arrows.meta,fit,backgrounds}
\usepackage[hidelinks]{hyperref}

\newcommand{\dataset}{\textsc{ISOT/Kaggle}}
\newcommand{\Fone}{$F_1$}
\DeclareMathOperator*{\argmax}{arg\,max}

\begin{document}

\title{What Does 99\% Accuracy Measure?\\
A Reproducible Audit of Shortcut Learning\\
in a Widely Used Fake News Corpus}

\author{\IEEEauthorblockN{Yuvraj Verma}
\IEEEauthorblockA{Independent Researcher, India\\
ORCID: 0009-0004-2138-3159\\
Email: yuvrajverma282004@gmail.com\\
Code and data: \url{https://github.com/vermayuvraj/fake-news-detection}}}

\maketitle

\begin{abstract}
Text classifiers trained on the \dataset{} ``Fake and Real News'' corpus
routinely report accuracy and \Fone{} above~0.98, a level of performance that
sits uneasily beside the difficulty of assessing veracity. We ask what those
numbers actually measure. Treating a conventional TF--IDF and linear-classifier
pipeline as a measurement instrument rather than a contribution, we audit the
corpus along three leakage channels and two distribution-shift protocols, with
all code, seeds, and derived numbers released. Three findings emerge. First,
the benchmark is partly degenerate: a classifier that discards the article text
entirely and observes only the \texttt{subject} metadata field attains
\Fone{}~$=1.000$, because the topic labels of the two classes do not overlap.
Second, removing all three leakage channels---metadata, a newswire source tag
present in 99.2\% of real articles, and 6{,}251 duplicate documents that
contaminate 19.4\% of a naive test split---lowers \Fone{} by only 1.21 points
(0.9935 to 0.9814), because the residual signal is diffuse editorial style
rather than a small set of giveaway tokens: nine representation variants span
just 0.6 \Fone{} points, 543 labelled documents already reach \Fone{}~$=0.933$,
and deleting the 1{,}000 highest-weight unigrams still leaves~0.926. Third, that
style signal does not transfer. Under a topic-disjoint protocol, average
precision falls from 0.9995 to 0.9475 and deployed \Fone{} from 0.9905
to~0.8067; a prior-matched resampling analysis attributes most of the latter
collapse to class-prior shift and threshold miscalibration
(\Fone{}~$=0.9427$) while confirming a genuine 5.2-point loss of
discrimination. Temporal transfer within the corpus is by contrast nearly
lossless. A fine-tuned DistilBERT sharpens rather than resolves the problem: it
is stronger in-distribution (\Fone{}~$=0.9993$) yet degrades far more under
topic shift, losing 12.9 points of average precision against the linear model's
5.2 and reaching prior-matched \Fone{} of only~0.689, while its in-domain
validation \Fone{} of 0.9995 gives no warning of that collapse. The decisive
test is cross-corpus: transferred to the independently built LIAR benchmark, all
three models fall to near-chance ranking (ROC-AUC $0.54$--$0.57$) and none beats
a majority-class baseline. We conclude that within-corpus scores on this
benchmark quantify source and topic separability rather than veracity
assessment, that added capacity exploits the shortcut more efficiently instead
of avoiding it, and we recommend metadata-only,
small-sample, and topic-disjoint baselines as inexpensive diagnostics that any
future study on this corpus can report.
\end{abstract}

\begin{IEEEkeywords}
fake news detection, shortcut learning, dataset bias, data leakage,
distribution shift, text classification, TF--IDF, reproducibility,
evaluation methodology
\end{IEEEkeywords}

\section{Introduction}
\label{sec:intro}

False news propagates measurably faster and more widely than true news on
social platforms \cite{vosoughi2018spread}, and its documented reach during the
2016 United States election \cite{allcott2017social} motivated sustained
interest in automated detection. The task, however, is difficult in a way that
resists purely lexical solutions: deciding whether a claim is true generally
requires evidence external to the text
\cite{thorne2018fever,zhou2020survey}. It is
therefore notable that a large body of applied work reports near-perfect
performance on the task, with accuracies above~0.99 on public corpora
\cite{ahmed2017detection,ahmed2018detecting,khan2021benchmark}. Either the
problem is easier than its formulation suggests, or the reported numbers
measure something other than veracity assessment. This paper investigates the
second possibility for one specific, heavily used corpus.

Our object of study is the ``Fake and Real News'' dataset
\cite{bisaillon2020dataset}, a redistribution of the corpus introduced by
Ahmed et al. \cite{ahmed2017detection,ahmed2018detecting} and among the most
frequently used public resources for this task
\cite{khan2021benchmark}. It pairs
21{,}417 genuine articles drawn from a single newswire against 23{,}481
articles from outlets flagged by fact-checking organisations. That construction
is convenient and, as we show, consequential: the two classes differ
systematically in provenance and house style, not merely in truthfulness.

The phenomenon we investigate is \emph{shortcut learning}: a model attaining
high benchmark scores by exploiting surface regularities that correlate with
the label in the dataset but not in the underlying task
\cite{geirhos2020shortcut}. Shortcut learning is well documented in natural
language inference, where hypothesis-only baselines and annotation artifacts
account for much of apparent model competence
\cite{gururangan2018annotation,poliak2018hypothesis,mccoy2019right,niven2019probing},
and in computer vision, where dataset identity is itself predictable
\cite{torralba2011unbiased}. For fake news specifically, Bozarth and Budak
\cite{bozarth2020toward} document how evaluation choices drive reported
performance, and Schuster et al. \cite{schuster2020limitations} show that
stylometric cues fail when provenance and veracity are decoupled. What has been
missing, to the best of our knowledge, is a per-channel, fully reproducible
decomposition for this particular corpus, combined with a shift analysis that
separates calibration effects from genuine degradation.

We deliberately do not propose a new architecture. A stronger model would
confound the question: if the benchmark is degenerate, a higher score is
evidence about the benchmark, not the model. We instead fix a transparent,
fully interpretable pipeline (TF--IDF features with four linear classifiers)
and vary the \emph{data} and the \emph{evaluation protocol},
using the classifier as a probe. Linear models make this strategy viable
because every decision decomposes into per-token weights that can be inspected
and ablated directly.

\noindent\textbf{Contributions.}
\begin{enumerate}
  \item A reproducible audit protocol (Section~\ref{sec:method}) that isolates
    three leakage channels in the \dataset{} corpus and quantifies each
    independently, released as executable code with fixed seeds
    (Section~\ref{sec:setup}).
  \item Evidence that the benchmark is partly degenerate: a metadata-only
    classifier that never sees the article text attains \Fone{}~$=1.000$
    (Section~\ref{sec:results-ablation}).
  \item A demonstration that the residual post-mitigation signal is
    \emph{diffuse} rather than localised, established through representation
    ablations, a learning curve, and a top-$K$ feature-deletion probe
    (Section~\ref{sec:results-representation}).
  \item A prior-controlled distribution-shift analysis
    (Section~\ref{sec:results-shift}) that decomposes an 18.4-point deployed
    \Fone{} collapse into a class-prior/calibration component and a genuine
    5.2-point loss of average precision.
  \item A capacity control (Section~\ref{sec:results-transformer}) in which a
    fine-tuned DistilBERT, trained on identical documents and splits, is
    stronger in-distribution but loses 2.5 times more average precision under
    topic shift, evidence that the limitation is a property of the data rather
    than of model expressiveness.
  \item A cross-corpus test (Section~\ref{sec:results-crosscorpus}) showing that
    all three ISOT-trained models transfer to the LIAR benchmark at near-chance
    ranking, the decisive external check that in-corpus scores do not reflect a
    transferable notion of veracity.
  \item Concrete, low-cost diagnostics we recommend as standard practice for
    future work on this corpus (Section~\ref{sec:discussion}).
\end{enumerate}

\section{Related Work}
\label{sec:related}

\subsection{Fake News Detection and Its Benchmarks}
Surveys of the area distinguish content-based approaches from those exploiting
social context and external evidence
\cite{shu2017fake,zhou2020survey,oshikawa2020survey}. Purely content-based
detection is attractive operationally, since it needs only the article, but it
is theoretically limited, since textual form underdetermines factual accuracy.
Within that paradigm, P\'erez-Rosas et al. \cite{perezrosas2018automatic}
report linguistic-feature classifiers across several domains, while Horne and
Adal{\i} \cite{horne2017this} find that headlines alone carry substantial
discriminative signal---an observation our input-field ablation revisits in
Section~\ref{sec:results-representation}. Zellers et al.
\cite{zellers2019defending} extend the problem to machine-generated articles,
where provenance and veracity come apart by construction. Benchmarks reflect
this tension. LIAR \cite{wang2017liar} supplies short
PolitiFact-labelled claims with fine-grained veracity labels; FEVER
\cite{thorne2018fever} pairs claims with Wikipedia evidence, making the
evidence-retrieval step explicit; NELA-GT \cite{norregaard2019nela} provides
large multi-labelled article collections with source-level annotations. The
\dataset{} corpus \cite{ahmed2017detection,ahmed2018detecting,bisaillon2020dataset}
differs in an important respect: its two classes were collected from disjoint
sets of publishers. Labels therefore carry provenance information, a property
central to our analysis.

Reported results on the corpus are uniformly high. Ahmed et al.\ obtained
accuracies in the high nineties using $n$-gram features with linear models
\cite{ahmed2017detection,ahmed2018detecting}; the benchmark study of Khan et al.
\cite{khan2021benchmark} finds that simple models remain competitive with
neural alternatives across several fake-news corpora, including this one, and
that scores cluster near the ceiling. Transformer-based systems such as
FakeBERT \cite{kaliyar2021fakebert} report comparable figures. Our reading of
this pattern is that near-ceiling agreement across model families of very
different capacity is itself diagnostic: when a linear bag-of-words model and a
pretrained transformer perform indistinguishably, the discriminative
information is likely to be shallow.

\subsection{Shortcut Learning and Dataset Artifacts}
Geirhos et al. \cite{geirhos2020shortcut} formalise shortcut learning as
reliance on decision rules that succeed on benchmark data but fail under
distribution shift. In NLI, Gururangan et al. \cite{gururangan2018annotation}
and Poliak et al. \cite{poliak2018hypothesis} show that hypothesis-only models
substantially outperform chance, revealing annotation artifacts; McCoy et al.
\cite{mccoy2019right} and Niven and Kao \cite{niven2019probing} show that
apparently competent models rely on syntactic heuristics and spurious
statistical cues. Diagnostic methodologies developed in response include
contrast sets \cite{gardner2020evaluating} and behavioural testing
\cite{ribeiro2020beyond}. Our metadata-only and small-sample probes are direct
analogues of the hypothesis-only baseline: cheap experiments whose success
indicates that the benchmark, not the model, deserves scrutiny.

\subsection{Leakage and Evaluation Protocol}
Kaufman et al. \cite{kaufman2012leakage} give the canonical taxonomy of
leakage; Kapoor and Narayanan \cite{kapoor2023leakage} document its prevalence
and its role in irreproducible machine-learning claims across scientific
fields. Duplicate records spanning train and test are a recognised instance.
Gorman and Bedrick \cite{gorman2019we} show that conclusions drawn from a
single standard split can reverse under resampling, motivating our use of
cross-validation and bootstrap intervals alongside a fixed split. On protocol
design specifically, Bozarth and Budak \cite{bozarth2020toward} show that
performance rankings for fake-news classifiers depend strongly on evaluation
choices, and argue for source-aware splits. Schuster et al.
\cite{schuster2020limitations} demonstrate that stylometric detectors degrade
sharply once provenance is decoupled from veracity. Our topic-disjoint protocol
operationalises that concern for this corpus, and our prior-matched analysis
adds a component their setting does not isolate: how much of an observed
collapse is attributable to class-prior shift rather than to loss of
discriminative power.

\subsection{Positioning}
Relative to prior work, this paper contributes neither a new model nor a new
dataset. Its contribution is measurement: a per-channel quantification of
leakage in a specific widely used corpus, an explicit demonstration that the
surviving signal is distributed rather than concentrated, and a shift analysis
that separates two mechanisms usually reported as one number. Every figure in
the paper is regenerated by a released script from the raw corpus.

\section{Problem Statement}
\label{sec:problem}

Let $\mathcal{D}=\{(d_i,y_i)\}_{i=1}^{N}$ be a corpus of news documents
$d_i \in \mathcal{X}$ with labels $y_i \in \{0,1\}$, where $y=1$ denotes
\emph{fake}. The nominal task is to learn $f:\mathcal{X}\rightarrow\{0,1\}$
minimising expected risk under the deployment distribution $\mathcal{P}_{\text{dep}}$:
\begin{equation}
  R_{\text{dep}}(f) \;=\; \mathbb{E}_{(d,y)\sim\mathcal{P}_{\text{dep}}}
  \big[\mathbb{I}[f(d)\neq y]\big].
  \label{eq:risk}
\end{equation}
Benchmark practice instead reports empirical risk on a held-out split of the
same corpus, i.e.\ an estimate of $R_{\text{bench}}(f)$ under the corpus
distribution $\mathcal{P}_{\text{bench}}$. The two coincide only if
$\mathcal{P}_{\text{bench}}\approx\mathcal{P}_{\text{dep}}$.

We formalise the failure mode of interest as follows. Write each document as a
pair $d=(s,c)$ where $c$ is veracity-bearing content and $s$ is a
\emph{provenance signature}: stylistic, formatting, and editorial regularities
determined by the publishing organisation. In $\mathcal{P}_{\text{bench}}$ the
two classes were sampled from disjoint publisher sets, so $s$ and $y$ are
strongly dependent:
\begin{equation}
  I(s;y)\Big|_{\mathcal{P}_{\text{bench}}} \;\gg\;
  I(s;y)\Big|_{\mathcal{P}_{\text{dep}}},
  \label{eq:mi}
\end{equation}
where $I(\cdot;\cdot)$ denotes mutual information. A learner minimising
empirical risk has no incentive to prefer $c$ over $s$; if $s$ is more easily
extracted, the learned rule will exploit it. Benchmark risk is then a biased
estimate of deployment risk, and the bias is not removed by increasing model
capacity or dataset size---only by changing the data or the protocol.

This yields three empirical questions, which organise our experiments:
\begin{description}
  \item[Q1] \emph{Explicit leakage.} How much of the benchmark score is
    attributable to identifiable artifacts (metadata fields, source tags, and
    duplicate documents) that can be removed by preprocessing?
  \item[Q2] \emph{Residual signal structure.} After removing those artifacts,
    is the remaining signal concentrated in a few tokens (hence removable) or
    diffusely distributed across editorial style (hence not)?
  \item[Q3] \emph{Transfer.} Does the surviving decision rule remain valid when
    the topic mix or time period changes, and how much of any observed
    degradation reflects loss of discrimination as opposed to class-prior
    shift?
  \item[Q4] \emph{Capacity.} Is the limitation a property of the instrument or
    of the data? If a pretrained contextual model, which can represent
    veracity-bearing content far better than a bag of words, nonetheless
    degrades at least as much under shift, the constraint lies in
    $\mathcal{P}_{\text{bench}}$ rather than in model expressiveness.
  \item[Q5] \emph{External validity.} Does anything learned on this corpus
    transfer to an independently constructed one? A model trained here and
    tested on a different benchmark isolates whatever veracity signal, if any,
    is not tied to this corpus's provenance structure.
\end{description}

\section{Proposed Methodology}
\label{sec:method}

\subsection{Design Rationale}
The instrument must be transparent, cheap, and deterministic, since it is
applied across 20 experimental conditions. We therefore use sparse TF--IDF
features with linear classifiers. This choice is deliberate on three grounds.
First, interpretability: a linear model assigns one scalar weight per token, so
the top-$K$ deletion probe of Section~\ref{sec:results-representation} has a
well-defined meaning \cite{ribeiro2016why}. Second, determinism: with a fixed seed and a
coordinate-descent solver the pipeline is bit-reproducible, which matters when
differences of interest are on the order of $10^{-2}$. Third, capacity control:
if a low-capacity model already saturates the benchmark, high scores cannot be
attributed to sophisticated inference. The trade-off is accepted knowingly: no modelling of word
order beyond short $n$-grams, and no lexical generalisation. and Section~\ref{sec:limitations} states where it constrains our
conclusions.

Distributed word representations \cite{mikolov2013efficient,pennington2014glove},
subword linear models \cite{joulin2017bag}, character-level convolutional
networks \cite{zhang2015character}, and pretrained transformers
\cite{vaswani2017attention,devlin2019bert} are all defensible alternatives, and
we return to them in Section~\ref{sec:futurework}. None is preferable
\emph{as an instrument} for the present question, because each replaces
inspectable per-token weights with distributed parameters that the deletion
probe of Section~\ref{sec:results-representation} could not target. Sparse
lexical features with linear models also remain strong baselines for topical
text classification \cite{joachims1998text,wang2012baselines}, so little
discriminative power is sacrificed.

\subsection{Corpus Construction and Leakage Control}
\label{sec:method-clean}
Algorithm~\ref{alg:corpus} specifies corpus construction. Three controls are
applied, each independently switchable so that its contribution can be
measured:

\textbf{(C1) Metadata exclusion.} The \texttt{subject} and \texttt{date} fields
are discarded. In this corpus the real class carries only the subjects
\texttt{politicsNews} and \texttt{worldnews} while the fake class carries six
entirely different values (Table~\ref{tab:corpus}); the supports are disjoint,
so \texttt{subject} alone determines the label. The \texttt{date} fields are
also not exchangeable: 2{,}926 fake articles fall outside the date range spanned
by the real class.

\textbf{(C2) Source-tag removal.} Real articles are newswire copy that
typically opens with a dateline of the form \texttt{CITY (Reuters) -- }; the
agency name appears in 99.21\% of real articles and 0.04\% of fake ones. We
strip the leading dateline and every remaining occurrence of the agency name.
The transformation is applied symmetrically to both classes, so it introduces
no new asymmetry.

\textbf{(C3) De-duplication.} Exact duplicates of the cleaned document string
are collapsed to a single instance \emph{before} splitting. Under a naive
protocol that omits this step, 19.37\% of test documents also occur verbatim in
the training set (Section~\ref{sec:results-ablation}), so test performance
partly measures memorisation.

Remaining normalisation is deliberately conservative and purely regular: the
title and body are concatenated, case is folded, URLs and e-mail addresses are
removed, characters outside \texttt{[a--z']} are mapped to whitespace, runs of
whitespace are collapsed, and documents shorter than 20 characters are dropped.
We do not stem or lemmatise. Stemming would add a dependency on an external
resource without addressing any of Q1--Q3, and with bigrams and stop-word
removal already in place its effect on this corpus is marginal.

\begin{algorithm}[t]
\caption{Leakage-controlled corpus construction}
\label{alg:corpus}
\begin{algorithmic}[1]
\Require Raw tables $T_{\text{real}}, T_{\text{fake}}$; flags
  $\textsc{c1},\textsc{c2},\textsc{c3}$; seed $\sigma$
\Ensure Disjoint splits $\mathcal{S}_{\text{tr}},\mathcal{S}_{\text{va}},\mathcal{S}_{\text{te}}$
\State $\mathcal{D} \gets$ concat$(T_{\text{real}}\!\times\!\{0\},\;T_{\text{fake}}\!\times\!\{1\})$
\ForAll{$(d,y) \in \mathcal{D}$}
  \State $u \gets \textsc{title}(d) \Vert \text{`` ''} \Vert \textsc{body}(d)$
  \If{$\neg\textsc{c1}$} \Comment{leakage condition only}
    \State $u \gets \textsc{subject}(d) \Vert \text{`` ''} \Vert u$
  \EndIf
  \State $u \gets \textsc{lowercase}(u)$
  \If{\textsc{c2}}
    \State $u \gets \textsc{stripDateline}(u)$;\;
           $u \gets \textsc{removeAgencyName}(u)$
  \EndIf
  \State $u \gets \textsc{removeUrlsEmails}(u)$;\;
         $u \gets \textsc{keepAlpha}(u)$
  \State $u \gets \textsc{collapseSpace}(u)$
\EndFor
\State $\mathcal{D} \gets \{(u,y) \in \mathcal{D} : |u| \geq 20\}$
\If{\textsc{c3}}
  \State $\mathcal{D} \gets \textsc{dropDuplicates}(\mathcal{D}, \text{key}=u)$
\EndIf
\State $(\mathcal{D}', \mathcal{S}_{\text{te}}) \gets
  \textsc{stratifiedSplit}(\mathcal{D}, 0.20, \sigma)$
\State $(\mathcal{S}_{\text{tr}}, \mathcal{S}_{\text{va}}) \gets
  \textsc{stratifiedSplit}(\mathcal{D}', 0.125, \sigma)$
  \Comment{$0.10/0.80$}
\State \Return $\mathcal{S}_{\text{tr}},\mathcal{S}_{\text{va}},\mathcal{S}_{\text{te}}$
\end{algorithmic}
\end{algorithm}

\subsection{Feature Representation}
Documents are mapped to sparse TF--IDF vectors
\cite{sparckjones1972statistical,salton1988term} over word unigrams and
bigrams. With $N$ training documents and $\mathrm{df}(t)$ the document
frequency of term $t$, we use sublinear term frequency and smoothed inverse
document frequency,
\begin{align}
  \mathrm{tf}'(t,d) &= 1 + \ln \mathrm{tf}(t,d), \label{eq:tf}\\
  \mathrm{idf}(t)   &= \ln\!\frac{1+N}{1+\mathrm{df}(t)} + 1,
  \label{eq:idf}
\end{align}
followed by $\ell_2$ normalisation of each document vector,
\begin{equation}
  x_{d,t} \;=\;
  \frac{\mathrm{tf}'(t,d)\,\mathrm{idf}(t)}
       {\big\lVert \mathrm{tf}'(\cdot,d)\odot\mathrm{idf}(\cdot)\big\rVert_2}.
  \label{eq:tfidf}
\end{equation}
Sublinear scaling prevents a term repeated many times in one document from
dominating; $\ell_2$ normalisation removes document-length effects, which
matters here because the classes differ in mean length (436.7 versus 395.8
tokens, Table~\ref{tab:corpus}). Terms occurring in fewer than five documents
or more than 90\% of documents are discarded, English stop words are removed,
and the vocabulary is capped at $50{,}000$ features by corpus frequency.

Critically, the vectoriser is fitted on the training split only and applied
unchanged to validation and test. Fitting on the full corpus would leak
document-frequency statistics of the evaluation data into the representation---a
subtle but real instance of the leakage this paper studies.

\subsection{Classifiers}
Four linear classifiers are trained on identical features. All four are
standard; we summarise them to fix notation and to make the interpretability
argument precise. Let $x\in\mathbb{R}^{V}$ be a document vector,
$w\in\mathbb{R}^{V}$ a weight vector, and $b$ a bias.

\emph{Logistic regression} models
$P(y{=}1\mid x)=\sigma(w^{\!\top}x+b)$ with
$\sigma(z)=(1+e^{-z})^{-1}$, fitted by $\ell_2$-regularised maximum likelihood.
With labels $\tilde{y}\in\{-1,+1\}$ the objective solved by the coordinate-descent
solver \cite{fan2008liblinear} is
\begin{equation}
  \min_{w,b}\;\tfrac{1}{2}\lVert w\rVert_2^2
  + C\sum_{i=1}^{n}\ln\!\big(1+e^{-\tilde{y}_i(w^{\!\top}x_i+b)}\big).
  \label{eq:lr}
\end{equation}
It is our reference probe because $w$ is directly interpretable and because
$\sigma(\cdot)$ yields calibrated scores.

\emph{Multinomial naive Bayes} \cite{mccallum1998comparison} estimates
per-class term distributions with additive smoothing $\alpha$,
\begin{equation}
\begin{split}
  \hat{P}(t\mid y) &= \frac{N_{y,t}+\alpha}{N_{y}+\alpha|V|},\\[2pt]
  \hat{y} &= \argmax_{y}\Big[\ln \hat{P}(y)
             + \sum_{t} x_{t}\ln\hat{P}(t\mid y)\Big],
\end{split}
  \label{eq:nb}
\end{equation}
under the assumption that terms are conditionally independent given the class.
The assumption is violated by construction in text---overlapping unigrams and
bigrams are strongly dependent---and we use the model as a deliberately
mis-specified reference point \cite{domingos1997optimality,rennie2003tackling}.

\emph{Linear support vector classification} \cite{cortes1995support,joachims1998text}
maximises the margin under squared hinge loss,
\begin{equation}
  \min_{w,b}\;\tfrac{1}{2}\lVert w\rVert_2^2
  + C\sum_{i=1}^{n}\max\!\big(0,\,1-\tilde{y}_i(w^{\!\top}x_i+b)\big)^{2}.
  \label{eq:svm}
\end{equation}

\emph{The passive--aggressive classifier} \cite{crammer2006online} is an online
large-margin learner. On example $(x_t,\tilde{y}_t)$ with hinge loss
$\ell_t=\max(0,1-\tilde{y}_t\,w_t^{\!\top}x_t)$, the PA-I update is
\begin{equation}
  \tau_t=\min\!\Big(C,\;\frac{\ell_t}{\lVert x_t\rVert_2^{2}}\Big),
  \qquad
  w_{t+1}=w_t+\tau_t\,\tilde{y}_t\,x_t,
  \label{eq:pa}
\end{equation}
i.e.\ the weights are left unchanged when the margin is already satisfied
($\ell_t=0$) and otherwise moved by the smallest step that corrects the current
example, with $C$ bounding the step. In all four models $C$ is an inverse
regularisation strength: larger $C$ penalises weight magnitude less.

\subsection{Evaluation Protocols}
\label{sec:method-protocols}
Deliberately shifting the test distribution is standard practice for exposing
brittle decision rules \cite{quinonero2009dataset,koh2021wilds}. Three
protocols partition the same cleaned corpus.

\textbf{Random (standard).} Stratified $70/10/20$ train/validation/test. This
reproduces conventional practice and provides the reference score.

\textbf{Topic-disjoint.} Training and test pools are drawn from disjoint
\texttt{subject} values: \texttt{politicsNews} (real) and
\texttt{News}, \texttt{politics}, \texttt{left-news} (fake) for training;
\texttt{worldnews} (real) and \texttt{Government News}, \texttt{US\_News},
\texttt{Middle-east} (fake) for test. Validation is carved from the training
pool only, so no target-domain labels are available at model-selection
time---the realistic deployment condition. Because subject values also proxy
for topic and sub-publication, this protocol perturbs provenance and topic
jointly.

\textbf{Temporal.} Documents are ordered by publication date and split
chronologically $70/10/20$, restricted to the window in which both classes are
present so that the test period is not trivially single-class.

Both shift protocols alter the class prior as a side effect
(Table~\ref{tab:protocols}). Since \Fone{} depends on the prior, a raw \Fone{}
comparison across protocols conflates prior shift with loss of discrimination.
Algorithm~\ref{alg:shift} therefore reports four complementary quantities:
\Fone{} at the default threshold (what a deployed system would achieve),
threshold-oracle \Fone{} (the best any cutoff on the model's own scores could
achieve, isolating miscalibration), average precision (threshold-free ranking
quality), and \Fone{} on prior-matched subsamples in which negatives are
resampled so that the fake ratio equals that of the random split. Only the last
two support a like-for-like comparison of discriminative power.

\begin{algorithm}[t]
\caption{Prior-controlled shift evaluation}
\label{alg:shift}
\begin{algorithmic}[1]
\Require Fitted model $f$ with scores $s(\cdot)$; test set
  $\mathcal{S}_{\text{te}}$; reference prior $\pi^{\star}$; repeats $M$; seed $\sigma$
\State $\hat{y}\gets f(\mathcal{S}_{\text{te}})$;\;
       $s\gets s(\mathcal{S}_{\text{te}})$
\State $F_1^{\text{def}} \gets F_1(y,\hat{y})$
  \Comment{deployed operating point}
\State $F_1^{\text{orc}} \gets \max_{\theta} F_1\big(y,\mathbb{I}[s>\theta]\big)$
  \Comment{isolates calibration}
\State $\mathrm{AP} \gets \textsc{averagePrecision}(y,s)$
  \Comment{threshold-free}
\State $\mathcal{P}\gets\{i: y_i{=}1\}$;\;
       $\mathcal{N}\gets\{i: y_i{=}0\}$
\State $m \gets \mathrm{round}\big(|\mathcal{P}|(1-\pi^{\star})/\pi^{\star}\big)$
\For{$j \gets 1$ \textbf{to} $M$}
  \State $\mathcal{N}_j \sim \textsc{sampleWithoutReplacement}(\mathcal{N},m;\sigma)$
  \State $\phi_j \gets F_1\big(y_{\mathcal{P}\cup\mathcal{N}_j},
         \hat{y}_{\mathcal{P}\cup\mathcal{N}_j}\big)$
\EndFor
\State \Return $F_1^{\text{def}},\,F_1^{\text{orc}},\,\mathrm{AP},\,
  \mathrm{mean}(\phi),\,\mathrm{sd}(\phi)$
\end{algorithmic}
\end{algorithm}

\subsection{Capacity Control: Fine-Tuned DistilBERT}
\label{sec:method-transformer}
Answering Q4 requires a model of substantially greater capacity evaluated under
identical conditions. We fine-tune DistilBERT \cite{sanh2019distilbert}, a
six-layer distillation of BERT \cite{devlin2019bert} with roughly 66M
parameters, whose subword tokenisation and contextual self-attention
\cite{vaswani2017attention} can in principle represent the semantic content that
a bag of words cannot.

Comparability is the whole point of the experiment, so two things are held
fixed and one is deliberately allowed to differ. The document set and the split
assignment are \emph{identical}: we rebuild the primary frame preserving the
original row index, replicate each protocol exactly, and assert that the
resulting split cardinalities match those in Table~\ref{tab:protocols} before
training proceeds. The leakage controls C1--C3 are also identical.

What differs is surface normalisation. The linear pipeline maps every character
outside \texttt{[a--z']} to whitespace, which suits a bag-of-words model but
discards punctuation and casing that a subword transformer can exploit. Applying
that destructive normalisation would handicap the transformer for reasons
unrelated to Q4, so the transformer receives a minimally normalised variant of
the \emph{same} documents: the source tag, URLs, and e-mail addresses are
removed and whitespace is collapsed, but punctuation and digits are retained.
This asymmetry favours the transformer, which is the conservative direction for
our argument---we are testing whether extra capacity \emph{helps}, so giving it
the better input strengthens any negative result.

Training uses the recipe standard for this model class: batch size 16, learning
rate $2\times10^{-5}$ with linear warmup over the first 10\% of steps, two
epochs, mixed-precision arithmetic, and a maximum sequence length of 256 subword
tokens. No hyperparameter search was performed, for the same reason it was
omitted for the linear models. Model selection follows the same rule as
the linear models---the per-epoch checkpoint with the best \emph{validation}
\Fone{} is the one evaluated on test. Because the corpus mean of 415 words
exceeds 256 subword tokens, we additionally repeat the random protocol at a
512-token limit to confirm that truncation is not what determines the outcome.

\subsection{Metrics}
We report accuracy, and precision, recall, and \Fone{} for the positive
(fake) class:
\begin{equation}
  P=\frac{\mathrm{TP}}{\mathrm{TP}+\mathrm{FP}},\quad
  R=\frac{\mathrm{TP}}{\mathrm{TP}+\mathrm{FN}},\quad
  F_1=\frac{2PR}{P+R}.
  \label{eq:prf}
\end{equation}
Taking \emph{fake} as positive makes the error types operationally meaningful: a
false positive suppresses a legitimate article, a false negative lets
misinformation through. We additionally report ROC-AUC \cite{hanley1982meaning},
which equals the probability that a randomly chosen fake article receives a
higher score than a randomly chosen real one and is invariant to the decision
threshold, and average precision, which is more informative than ROC-AUC when
the positive class is rare---as it becomes under both shift protocols. Model
selection uses validation \Fone{} throughout, a rule fixed before any test set
was examined.

\section{System Architecture}
\label{sec:architecture}

\begin{figure*}[t]
\centering
\begin{tikzpicture}[
  node distance=6mm and 7mm,
  pbox/.style={draw, rounded corners=1.5pt, align=center, font=\scriptsize,
              minimum height=8.5mm, inner sep=2pt, fill=blue!4},
  cbox/.style={draw, rounded corners=1.5pt, align=center, font=\scriptsize,
              minimum height=8.5mm, inner sep=2pt, fill=red!6, dashed},
  obox/.style={draw, rounded corners=1.5pt, align=center, font=\scriptsize,
              minimum height=8.5mm, inner sep=2pt, fill=green!6},
  arw/.style={-{Stealth[length=1.6mm]}, thick}
]
\node[pbox] (raw) {Raw corpus\\\texttt{True.csv} 21{,}417\\\texttt{Fake.csv} 23{,}481};
\node[cbox, right=of raw] (ctl) {Leakage controls\\C1 metadata\\C2 source tag\\C3 duplicates};
\node[pbox, right=of ctl] (clean) {Normalise\\lowercase, URLs,\\alpha-only, collapse};
\node[pbox, right=of clean] (split) {Protocol split\\random /\\topic-disjoint /\\temporal};
\node[pbox, right=of split] (tfidf) {TF--IDF\\1--2 grams\\$V=50{,}000$\\\emph{fit on train}};
\node[pbox, right=of tfidf] (models) {Linear models\\LR, MNB,\\SVC, PA};
\node[obox, right=of models] (sel) {Select on\\validation \Fone{}};
\node[obox, below=8mm of sel] (test) {Single test\\evaluation};
\node[obox, left=of test] (an) {Analyses\\ablation, shift,\\deletion probe};
\node[obox, left=of an] (rep) {Reported\\tables and\\figures};

\draw[arw] (raw) -- (ctl);
\draw[arw] (ctl) -- (clean);
\draw[arw] (clean) -- (split);
\draw[arw] (split) -- (tfidf);
\draw[arw] (tfidf) -- (models);
\draw[arw] (models) -- (sel);
\draw[arw] (sel) -- (test);
\draw[arw] (test) -- (an);
\draw[arw] (an) -- (rep);
\draw[arw, dashed] (split.south) |- ++(0,-6mm) -| (an.north);
\node[font=\scriptsize, anchor=west] at ([yshift=-3.2mm]split.south west)
  {protocol varied};
\end{tikzpicture}
\caption{Audit architecture. The measurement pipeline (solid, blue) is held
fixed while the leakage controls C1--C3 (dashed, red) and the evaluation
protocol are varied. The validation split governs model selection; the test
split is evaluated exactly once per condition. Every reported figure is
regenerated from the raw corpus by the released scripts.}
\label{fig:architecture}
\end{figure*}
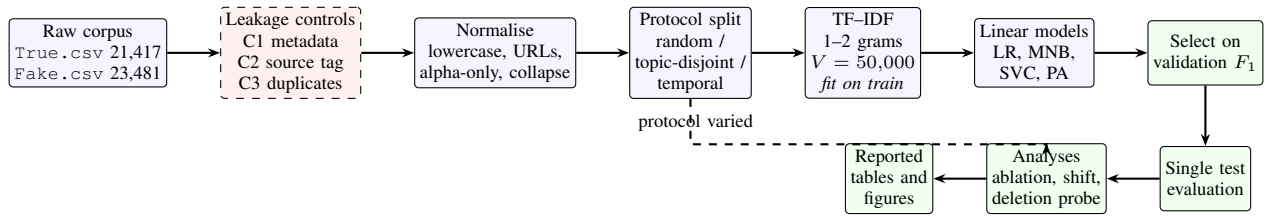

Figure~\ref{fig:architecture} shows the system. It is organised so that exactly
one factor varies per experimental condition: the preprocessing controls C1--C3
of Section~\ref{sec:method-clean}, the representation hyperparameters of
Equations~\eqref{eq:tf}--\eqref{eq:tfidf}, or the splitting protocol of
Section~\ref{sec:method-protocols}. Configuration is centralised in a single
module and serialised into the results file with every run, so each reported
number carries the exact settings that produced it.

\subsection{Computational Complexity}
Let $N$ be the number of training documents, $\bar{L}$ the mean token count,
$V$ the vocabulary size, and $z$ the mean number of non-zero features per
document. Vectorisation is $O(N\bar{L})$ time, dominated by tokenisation and
$n$-gram enumeration, with a vocabulary-construction pass over
$O(N\bar{L})$ candidate terms before frequency pruning. Training the linear
models costs $O(\kappa N z)$ for $\kappa$ passes: coordinate descent for
logistic regression and the SVM \cite{fan2008liblinear}, a single closed-form
counting pass for naive Bayes, and $\kappa$ online epochs for
passive--aggressive. Inference is $O(z)$ per document---one sparse dot product.

The measured quantities corroborate these rates and explain why the audit is
cheap enough to repeat across 20 conditions. On the training split
$z=190.1$ of $V=50{,}000$ features, i.e.\ a density of $0.38\%$; the sparse
matrix holds $5.17\times10^{6}$ non-zeros in $\approx62$\,MB at 12 bytes per
stored value, whereas a dense equivalent would require $\approx10.9$\,GB. Sparse
storage is thus not an optimisation but a precondition. Section~\ref{sec:results-cost}
reports wall-clock figures.

\section{Experimental Setup}
\label{sec:setup}

\textbf{Data.} The corpus is obtained from its public distribution
\cite{bisaillon2020dataset} and is not redistributed with our code.
Table~\ref{tab:corpus} reports its composition together with the leakage
indicators discussed in Section~\ref{sec:method-clean}; all values are computed
by the released script rather than quoted from prior work. The cross-corpus test
(Section~\ref{sec:results-crosscorpus}) additionally uses the LIAR test split
\cite{wang2017liar}: $1{,}267$ short political statements (mean $18.4$ tokens,
against ISOT's $415$) fact-checked by PolitiFact. LIAR's six-way labels are
binarised with the conventional cut---\{pants-fire, false, barely-true\}
$\rightarrow$ fake and \{half-true, mostly-true, true\} $\rightarrow$ real---which
yields a test prior of $0.436$ fake. The barely-true/half-true boundary is the
one debatable choice; moving it shifts the prior but not the near-chance result
of Section~\ref{sec:results-crosscorpus}.

\textbf{Hardware and software.} All experiments run on a single machine: Intel
Core i7-14700HX (28 logical cores), 15.7\,GB RAM, Windows~10 (AMD64). The linear
pipeline uses the CPU only; the DistilBERT control uses one NVIDIA GeForce
RTX~4050 Laptop GPU (6\,GB), peaking at 3.9\,GB of device memory. The stack is
Python~3.10.11 with scikit-learn~1.7.2 \cite{pedregosa2011scikit},
NumPy~1.24.3 \cite{harris2020array}, SciPy~1.15.3 \cite{virtanen2020scipy},
pandas~2.3.3 \cite{mckinney2010data}, and Matplotlib~3.10.8
\cite{hunter2007matplotlib}; the transformer control adds PyTorch~2.6.0 (CUDA
12.4) and Transformers~4.56.1. Exact versions are pinned in the released
\texttt{requirements.txt}.

\textbf{Reproducibility.} A single seed ($\sigma=42$) governs the splits and all
stochastic model components. Text normalisation is purely regular; the
vocabulary is frequency-ordered and deterministic; logistic regression uses the
deterministic \texttt{liblinear} solver. Two independent executions of the
released pipeline produced bit-identical metrics. This guarantee covers the
linear pipeline; the DistilBERT control is seeded identically but runs
mixed-precision CUDA kernels whose reduction order is not guaranteed, so its
figures should be expected to reproduce to roughly three decimal places rather
than exactly. Following community
recommendations on reporting practice
\cite{pineau2021improving,reimers2017reporting} we release code, seeds,
configuration snapshots, and the machine-readable results file from which every
table and figure in this paper is generated. Appendix~\ref{app:checklist}
contains the reproducibility checklist and Appendix~\ref{app:hyper} the full
hyperparameter listing.

\textbf{Statistical procedure.} Point estimates on the fixed test split are
accompanied by percentile bootstrap 95\% confidence intervals over $2000$
resamples of the test set \cite{efron1993introduction}. Because conclusions from
a single split can be fragile \cite{gorman2019we}, we additionally report
five-fold stratified cross-validation \Fone{} on the union of the training and
validation splits, refitting the vectoriser within each fold. Pairwise model
comparisons use McNemar's exact test on discordant predictions
\cite{mcnemar1947note,dietterich1998approximate}, the appropriate paired test
when two classifiers are evaluated on the same test set
\cite{dror2018hitchhiker}; omnibus procedures designed for comparisons across
many datasets \cite{demsar2006statistical} do not apply to a single-corpus
design. Prior-matched \Fone{} is averaged over $M=20$ resamples
(Algorithm~\ref{alg:shift}).

\textbf{Sanity control.} As a check that the harness itself introduces no
leakage, we ran the complete primary pipeline with the labels randomly
permuted. Test ROC-AUC was $0.4974$, statistically indistinguishable from the
chance value of $0.5$, confirming that the reported performance originates in
the data rather than in an implementation defect.

\begin{table}[t]
\centering
\caption{Corpus composition and leakage indicators. All quantities are computed
from the raw distribution by the released script.}
\label{tab:corpus}
\footnotesize
\begin{tabular}{@{}lrr@{}}
\toprule
Property & Real ($y{=}0$) & Fake ($y{=}1$) \\
\midrule
Articles (raw) & 21{,}417 & 23{,}481 \\
\midrule
\multicolumn{3}{@{}l}{\emph{\texttt{subject} values (disjoint supports)}}\\
\quad \texttt{politicsNews} / \texttt{worldnews} & 11{,}272 / 10{,}145 & -- \\
\quad \texttt{News} / \texttt{politics} & -- & 9{,}050 / 6{,}841 \\
\quad \texttt{left-news} / \texttt{Government News} & -- & 4{,}459 / 1{,}570 \\
\quad \texttt{US\_News} / \texttt{Middle-east} & -- & 783 / 778 \\
\midrule
Contains agency name ``(Reuters)'' & 99.21\% & 0.04\% \\
Duplicate body strings & 225 & 6{,}026 \\
Bodies shorter than 50 characters & 1 & 835 \\
Mean tokens per document (cleaned) & 395.8 & 436.7 \\
\midrule
\multicolumn{3}{@{}l}{\emph{After normalisation and de-duplication}}\\
\quad Documents retained & \multicolumn{2}{r}{38{,}826 of 44{,}898} \\
\quad Fake proportion & \multicolumn{2}{r}{0.461} \\
\quad Date span & \multicolumn{2}{r}{2015-03-31 to 2018-02-19} \\
\bottomrule
\end{tabular}
\end{table}

\begin{table}[t]
\centering
\caption{Evaluation protocols. Both shift protocols alter the class prior as a
side effect, which motivates the prior-controlled analysis of
Algorithm~\ref{alg:shift}.}
\label{tab:protocols}
\footnotesize
\begin{tabular}{@{}lrrrr@{}}
\toprule
Protocol & Train & Val. & Test & Test $\pi_{\text{fake}}$\\
\midrule
Random (standard) & 27{,}177 & 3{,}883 & 7{,}766 & 0.461\\
Topic-disjoint    & 24{,}326 & 3{,}476 & 11{,}024 & 0.119\\
Temporal          & 25{,}758 & 3{,}679 & 7{,}361 & 0.113\\
\bottomrule
\end{tabular}
\end{table}

\section{Results}
\label{sec:results}

\subsection{Reference Performance}
\label{sec:results-primary}
\begin{figure}[t]
\centering
\includegraphics[width=0.62\columnwidth]{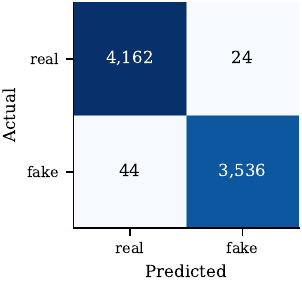}
\caption{Confusion matrix of the selected passive--aggressive model on the
random-protocol test split ($n=7{,}766$): $3{,}536$ true positives, $4{,}162$
true negatives, 24 false positives, 44 false negatives. The 24 false positives
are legitimate articles that would be suppressed; the 44 false negatives are
fake articles that would pass.}
\label{fig:confusion}
\end{figure}

Table~\ref{tab:primary} reports the four classifiers under the standard random
protocol. Passive--aggressive attains the best validation \Fone{} ($0.9924$)
and is selected; on the test split it reaches accuracy $0.9912$, precision
$0.9933$, recall $0.9877$, \Fone{} $0.9905$, and ROC-AUC $0.9995$, reproducing
the range reported in the literature for this corpus
\cite{ahmed2018detecting,khan2021benchmark}. Its confusion matrix
(Figure~\ref{fig:confusion}) contains 24 false positives and 44 false
negatives out of $7{,}766$ test documents. Bootstrap intervals are tight---\Fone{}
$0.9905$, 95\% CI $[0.9883, 0.9928]$---and five-fold cross-validation agrees
closely ($0.9916 \pm 0.0015$), so the reference figure is not an artefact of the
particular split.

Two observations already qualify the headline number. First, the three
margin- and likelihood-based models are separated by very little: McNemar's
exact test finds no significant difference between passive--aggressive and
LinearSVC ($n_{01}{=}8$, $n_{10}{=}11$, $p=0.65$), so the selected ``winner''
is statistically indistinguishable from the runner-up, and reporting it as the
best model would overstate what the data support. Differences against logistic
regression ($p=3.2\times10^{-12}$) and naive Bayes ($p=3.5\times10^{-55}$) are
significant. Second, the mis-specified naive Bayes baseline still reaches
\Fone{}~$0.9558$. When a model whose independence assumption is violated by
construction lands within four points of the best system, the task as posed is
unlikely to require sophisticated inference.

\begin{table}[t]
\centering
\caption{Reference performance under the random protocol. Selection uses
validation \Fone{}; test columns are computed once for the selected model and,
for completeness, for the remaining models. CV is five-fold stratified \Fone{}
on train${\,\cup\,}$validation. The final column gives McNemar $p$-values
against the selected model.}
\label{tab:primary}
\scriptsize
\setlength{\tabcolsep}{2.1pt}
\begin{tabular}{@{}lccccccc@{}}
\toprule
& \multicolumn{1}{c}{Val.} & \multicolumn{5}{c}{Test} & \\
\cmidrule(lr){2-2}\cmidrule(lr){3-7}
Model & \Fone{} & Acc. & $P$ & $R$ & \Fone{} & AUC & CV \Fone{} \\
\midrule
Logistic regression & 0.9840 & 0.9830 & 0.9906 & 0.9723 & 0.9814 & 0.9987 & $0.9829${\scriptsize$\pm.0030$}\\
Multinomial NB      & 0.9569 & 0.9592 & 0.9538 & 0.9578 & 0.9558 & 0.9911 & $0.9570${\scriptsize$\pm.0010$}\\
LinearSVC           & 0.9922 & 0.9909 & 0.9933 & 0.9869 & 0.9901 & 0.9995 & $0.9906${\scriptsize$\pm.0015$}\\
\textbf{Passive--aggressive}$^{\dagger}$ & \textbf{0.9924} & \textbf{0.9912} & \textbf{0.9933} & \textbf{0.9877} & \textbf{0.9905} & \textbf{0.9995} & $\mathbf{0.9916}${\scriptsize$\pm.0015$}\\
\midrule
\multicolumn{8}{@{}l}{\footnotesize $^{\dagger}$selected. McNemar vs.\ selected:
LinearSVC $p{=}0.65$ (n.s.);}\\
\multicolumn{8}{@{}l}{\footnotesize logistic regression
$p{=}3.2{\times}10^{-12}$; naive Bayes $p{=}3.5{\times}10^{-55}$.}\\
\multicolumn{8}{@{}l}{\footnotesize Bootstrap 95\% CI (selected):
Acc.\ $[.9892,.9933]$, \Fone{} $[.9883,.9928]$,}\\
\multicolumn{8}{@{}l}{\footnotesize $P$ $[.9905,.9958]$, $R$ $[.9839,.9913]$,
AUC $[.9992,.9997]$.}\\
\bottomrule
\end{tabular}
\end{table}

\subsection{Q1: Explicit Leakage Channels}
\label{sec:results-ablation}
\begin{figure}[t]
\centering
\includegraphics[width=\columnwidth]{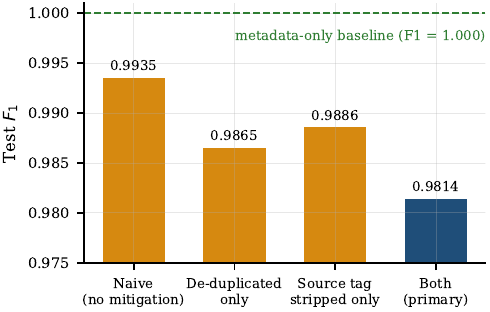}
\caption{Effect of the leakage mitigations on test \Fone{} (logistic-regression
probe). Removing all three channels costs $1.21$ points, leaving performance
close to the ceiling. The dashed line marks the metadata-only baseline, which
solves the benchmark exactly without reading any article text.}
\label{fig:ablation}
\end{figure}

Table~\ref{tab:ablation} and Figure~\ref{fig:ablation} isolate the three
channels using logistic regression as a fixed probe.

The most consequential result requires no classifier of the article at all. A
model whose only input is the \texttt{subject} field, with the article text
discarded entirely, attains \Fone{}~$=1.000$ and accuracy~$=1.000$ on a
held-out split, with a vocabulary of nine tokens. This is not a subtle
correlation but a deterministic mapping: the two classes have disjoint subject
supports (Table~\ref{tab:corpus}), so any model given this field can recover
the label exactly. Any experiment on this corpus that retains \texttt{subject}
among its features is measuring nothing about text.

Restricting inputs to title and body, the naive configuration reaches
\Fone{}~$0.9935$. De-duplication alone costs $0.70$ points and source-tag
removal alone $0.49$ points; applied together they cost $1.21$ points, yielding
the primary figure of $0.9814$. The duplicate channel has a clear mechanism:
under the naive protocol $1{,}739$ of $8{,}978$ test documents ($19.37\%$) occur
verbatim in the training set, so roughly a fifth of the test score reflects
retrieval of memorised strings rather than generalisation.

The honest reading of Table~\ref{tab:ablation} is two-sided, and we state both
directions. Each channel is real, measurable, and worth removing; a study that
ignores all three overstates its result by more than a point and, if it retains
metadata, by far more. Yet the mitigations do \emph{not} bring performance down
to a plausible level for veracity assessment---$0.9814$ remains close to the
ceiling. Explicit artifacts are therefore not the principal explanation for the
benchmark's easiness, which motivates Q2.

\begin{table}[t]
\centering
\caption{Leakage ablation with a fixed logistic-regression probe. $\Delta$ is
the change in test \Fone{} relative to the naive configuration. The
metadata-only row uses the \texttt{subject} field alone, with the article text
discarded.}
\label{tab:ablation}
\scriptsize
\setlength{\tabcolsep}{2.6pt}
\begin{tabular}{@{}llcccc@{}}
\toprule
Configuration & Docs & Acc. & \Fone{} & AUC & $\Delta$\Fone{}\\
\midrule
Metadata only (\texttt{subject}) & 44{,}898 & 1.0000 & 1.0000 & 1.000 & $+0.0065$\\
\midrule
Naive (no mitigation)            & 44{,}889 & 0.9932 & 0.9935 & 0.9996 & --\\
\quad $+$ de-duplication (C3)    & 38{,}826 & 0.9876 & 0.9865 & 0.9994 & $-0.0070$\\
\quad $+$ source tag removed (C2)& 44{,}889 & 0.9881 & 0.9886 & 0.9992 & $-0.0049$\\
\quad $+$ both (primary)         & 38{,}826 & 0.9830 & 0.9814 & 0.9987 & $-0.0121$\\
\bottomrule
\end{tabular}
\end{table}

\subsection{Q2: The Residual Signal Is Diffuse}
\label{sec:results-representation}
Three independent probes indicate that what survives mitigation is not a small
set of removable cues but a pervasive stylistic difference.

\textbf{Representation is nearly irrelevant.} Table~\ref{tab:representation}
varies the feature space over nine configurations spanning $n$-gram order,
vocabulary size across a 35-fold range, term weighting, and stop-word handling.
Test \Fone{} lies between $0.9773$ and $0.9835$---a spread of $0.62$ points.
A $5{,}000$-term unigram-and-bigram vocabulary ($0.9835$) slightly
\emph{outperforms} the $50{,}000$-term default and the $175{,}406$-term
unrestricted vocabulary ($0.9800$). When a tenfold reduction in capacity does
not degrade performance, the discriminative information is heavily redundant.

\begin{figure}[t]
\centering
\includegraphics[width=\columnwidth]{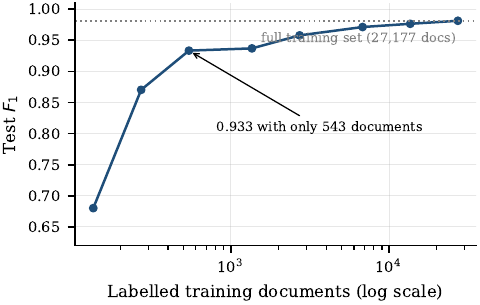}
\caption{Learning curve of the logistic-regression probe under the primary
configuration. Two per cent of the training data recovers $95.1\%$ of the
full-data \Fone{}, indicating a signal that is easy to acquire.}
\label{fig:learning}
\end{figure}

\textbf{Very little data suffices.} Figure~\ref{fig:learning} shows the
learning curve. With $543$ labelled documents, $2\%$ of the training
split, the probe reaches \Fone{}~$0.9334$, i.e.\ $95.1\%$ of the full-data
figure; $135$ documents already yield $0.6802$. Tasks requiring genuine
inference do not typically saturate on a few hundred examples.

\begin{figure}[t]
\centering
\includegraphics[width=\columnwidth]{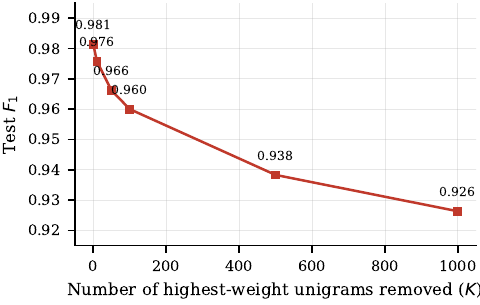}
\caption{Shortcut concentration. Deleting the $K$ highest-weight unigrams from
the vocabulary degrades \Fone{} only gradually; at $K=1000$ the model remains
within $5.5$ points of the unmodified baseline. The discriminative signal is
distributed across many stylistic features rather than concentrated in a few.}
\label{fig:deletion}
\end{figure}

\textbf{Deleting the strongest cues is not enough.} Ranking unigrams by
$|w|$ under the probe and removing the top $K$ from the vocabulary entirely
(so bigrams containing them are also destroyed) yields
Figure~\ref{fig:deletion}: \Fone{} declines gracefully from $0.9814$ at $K=0$
to $0.9600$ at $K=100$ and $0.9263$ at $K=1000$. Removing the thousand most
discriminative unigrams (among them \emph{said}, \emph{video},
\emph{washington}, \emph{image}, \emph{featured}, \emph{getty}, and the
weekday names) still leaves a model within $5.5$ points of the original. The
shortcut is not a token; it is a register.

\begin{figure}[t]
\centering
\includegraphics[width=\columnwidth]{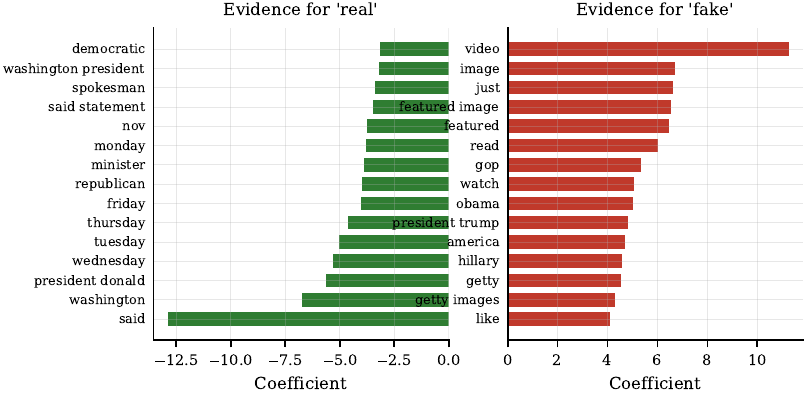}
\caption{Highest-magnitude logistic-regression coefficients under the primary
configuration. Real-class evidence reflects newswire house style; fake-class
evidence reflects web-publishing conventions such as embedded media and image
credits. Neither group encodes veracity.}
\label{fig:coefficients}
\end{figure}

Figure~\ref{fig:coefficients} makes that register legible. The strongest
evidence for \emph{real} consists of newswire conventions (\emph{said},
attribution phrasing, weekday datelines, \emph{factbox}); the strongest
evidence for \emph{fake} consists of web-publishing conventions
(\emph{video}, \emph{featured image}, \emph{getty images}, \emph{watch}).
These are properties of editorial workflow and content-management systems, not
of factual accuracy---exactly the provenance signature $s$ of
Section~\ref{sec:problem}.

\begin{table}[t]
\centering
\caption{Representation and input-field ablation (logistic-regression probe,
primary cleaning, random protocol). Performance is insensitive to the feature
space across a 35-fold range of vocabulary size.}
\label{tab:representation}
\footnotesize
\setlength{\tabcolsep}{4pt}
\begin{tabular}{@{}llcc@{}}
\toprule
Axis & Variant & Vocabulary & Test \Fone{}\\
\midrule
\multirow{3}{*}{$n$-gram order}
 & unigram            & 30{,}473 & 0.9801\\
 & unigram $+$ bigram (default) & 50{,}000 & 0.9814\\
 & $+$ trigram        & 50{,}000 & 0.9820\\
\midrule
\multirow{4}{*}{Vocabulary cap}
 & 5{,}000            & 5{,}000  & \textbf{0.9835}\\
 & 10{,}000           & 10{,}000 & 0.9823\\
 & 25{,}000           & 25{,}000 & 0.9817\\
 & unrestricted       & 175{,}406 & 0.9800\\
\midrule
\multirow{2}{*}{Weighting}
 & without sublinear tf & 50{,}000 & 0.9773\\
 & stop words retained  & 50{,}000 & 0.9828\\
\midrule
\multirow{3}{*}{Input field}
 & title only         & 39{,}615 & 0.9265\\
 & body only          & 49{,}997 & 0.9772\\
 & title $+$ body (default) & 50{,}000 & 0.9814\\
\bottomrule
\end{tabular}
\end{table}

\subsection{Q3: Transfer Under Distribution Shift}
\label{sec:results-shift}
\begin{figure}[t]
\centering
\includegraphics[width=\columnwidth]{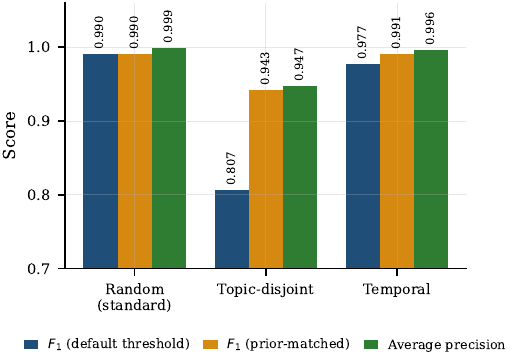}
\caption{Selected model across protocols. The gap between the two \Fone{} bars
under topic-disjoint evaluation is the contribution of class-prior shift and
threshold miscalibration; the drop in average precision is the genuine loss of
discriminative power. Temporal shift leaves both essentially intact.}
\label{fig:shift}
\end{figure}

\begin{figure}[t]
\centering
\includegraphics[width=\columnwidth]{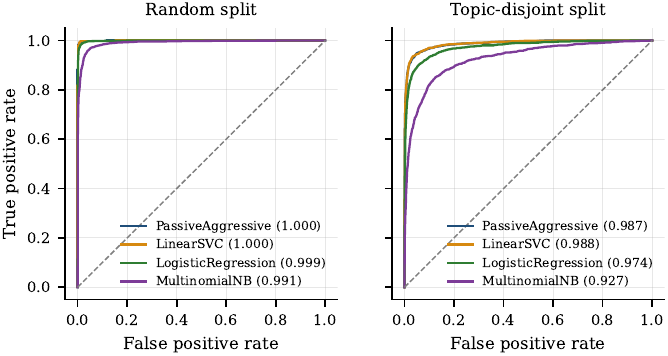}
\caption{ROC curves under the random (left) and topic-disjoint (right)
protocols, with ROC-AUC in parentheses. ROC-AUC remains high under topic shift
even where deployed \Fone{} collapses, illustrating why threshold-free ranking
metrics alone can mask an unusable operating point when the positive class
becomes rare.}
\label{fig:roc}
\end{figure}

Table~\ref{tab:shift} and Figure~\ref{fig:shift} report the prior-controlled
analysis; Figure~\ref{fig:roc} contrasts ROC behaviour between protocols.

Under the topic-disjoint protocol the deployed operating point degrades
severely: \Fone{} for the selected model falls from $0.9905$ to $0.8067$
($-18.4$ points), and for logistic regression from $0.9814$ to $0.6996$
($-28.2$ points). Precision drives the collapse (passive--aggressive:
$0.9933 \rightarrow 0.7011$) while recall is largely preserved
($0.9877 \rightarrow 0.9498$), the signature of a decision threshold that is no
longer appropriate for the target class prior, which falls from $0.461$ to
$0.119$.

Decomposing that collapse is essential for an honest claim, and it cuts both
ways. Matching the prior by resampling negatives recovers most of the loss
(\Fone{}~$0.9427 \pm 0.0032$), and the threshold-oracle value ($0.8887$) shows
that a substantial part of the remainder is miscalibration rather than an
inability to rank. We therefore do \emph{not} claim an 18-point loss of
discriminative power. However, average precision---which is threshold-free and
so immune to both effects---falls from $0.9995$ to $0.9475$, a genuine $5.2$-point
degradation, and balanced accuracy falls from $0.9910$ to $0.9475$. Logistic
regression degrades further (AP $0.9985 \rightarrow 0.9018$, $-9.7$ points),
indicating that the more heavily regularised probe leans harder on cues that do
not transfer. The correct statement is thus: under topic shift the model retains
much of its ranking ability but loses a real and measurable fraction of it, and
its deployed decision rule becomes unusable without recalibration on target
data that a real deployment would not possess.

Temporal shift, by contrast, is nearly benign: average precision moves only
from $0.9995$ to $0.9964$ and prior-matched \Fone{} is statistically
indistinguishable from the random-protocol value
($0.9914 \pm 0.0013$ versus $0.9905$). The asymmetry is informative. Publisher
house style is stable over the 2016--2017 window, so a chronological split, the
protocol often recommended as a realism improvement, does \emph{not} expose
the shortcut. Only perturbing provenance and topic does. Practitioners seeking a
stress test for this corpus should therefore prefer topic- or source-disjoint
splits over temporal ones.

\begin{table*}[t]
\centering
\caption{Prior-controlled distribution-shift analysis. $F_1^{\text{def}}$ is
the deployed operating point; $F_1^{\text{orc}}$ is the best \Fone{} attainable
by any threshold on the model's own scores; AP is average precision;
$F_1^{\text{match}}$ is \Fone{} on subsamples whose class prior is resampled to
that of the random protocol (mean $\pm$ s.d.\ over $M{=}20$ draws). Comparing
$F_1^{\text{def}}$ across protocols conflates prior shift with loss of
discrimination; AP and $F_1^{\text{match}}$ do not.}
\label{tab:shift}
\footnotesize
\begin{tabular}{@{}llccccccc@{}}
\toprule
Protocol & Model & $\pi_{\text{fake}}$ & $P$ & $R$ &
$F_1^{\text{def}}$ & $F_1^{\text{orc}}$ & AP & $F_1^{\text{match}}$\\
\midrule
\multirow{3}{*}{Random (standard)}
 & Logistic regression & 0.461 & 0.9906 & 0.9723 & 0.9814 & 0.9839 & 0.9985 & $0.9814$\\
 & LinearSVC           & 0.461 & 0.9933 & 0.9869 & 0.9901 & 0.9921 & 0.9995 & $0.9901$\\
 & Passive--aggressive & 0.461 & 0.9933 & 0.9877 & 0.9905 & 0.9918 & 0.9995 & $0.9905$\\
\midrule
\multirow{3}{*}{Topic-disjoint}
 & Logistic regression & 0.119 & 0.5596 & 0.9330 & 0.6996 & 0.8372 & 0.9018 & $0.9100${\scriptsize$\pm.0043$}\\
 & LinearSVC           & 0.119 & 0.7023 & 0.9460 & 0.8061 & 0.8886 & 0.9482 & $0.9408${\scriptsize$\pm.0036$}\\
 & Passive--aggressive & 0.119 & 0.7011 & 0.9498 & 0.8067 & 0.8887 & 0.9475 & $0.9427${\scriptsize$\pm.0032$}\\
\midrule
\multirow{3}{*}{Temporal}
 & Logistic regression & 0.113 & 0.9098 & 0.9723 & 0.9400 & 0.9577 & 0.9897 & $0.9783${\scriptsize$\pm.0018$}\\
 & LinearSVC           & 0.113 & 0.9681 & 0.9867 & 0.9773 & 0.9797 & 0.9963 & $0.9908${\scriptsize$\pm.0011$}\\
 & Passive--aggressive & 0.113 & 0.9670 & 0.9880 & 0.9774 & 0.9809 & 0.9964 & $0.9914${\scriptsize$\pm.0013$}\\
\bottomrule
\end{tabular}
\end{table*}

\subsection{Q4: Pretrained Capacity Amplifies the Shortcut}
\label{sec:results-transformer}
\begin{figure}[t]
\centering
\includegraphics[width=\columnwidth]{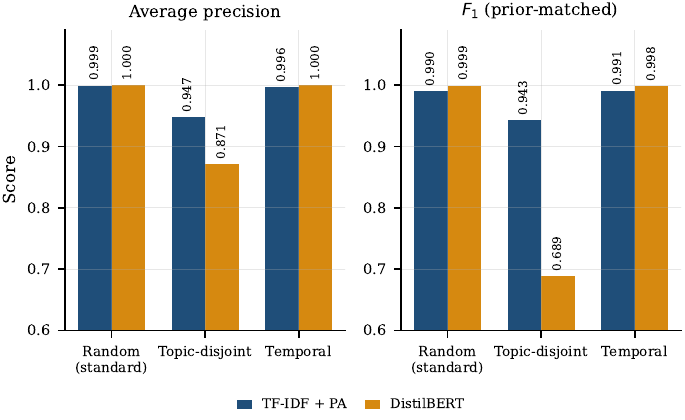}
\caption{Prior-robust comparison across protocols. DistilBERT matches or exceeds
the linear model in-distribution and under temporal shift, but loses
substantially more under topic shift on both average precision and
prior-matched \Fone{}. Greater capacity fits the provenance signature more
sharply rather than replacing it with veracity-bearing evidence.}
\label{fig:transformer}
\end{figure}

Table~\ref{tab:transformer} and Figure~\ref{fig:transformer} report the
DistilBERT control against the selected linear model on identical documents and
splits.

In-distribution the transformer is clearly the stronger classifier, as expected:
\Fone{} rises from $0.9905$ to $0.9993$ and average precision from $0.9995$ to
$1.0000$ (to four decimals; the unrounded value is $0.999999$). The random-split
benchmark therefore rewards capacity, and a study reporting only this row would
conclude that a pretrained transformer is the better system.

Under topic shift the ordering reverses sharply, and it does so on precisely the
metrics that are robust to the prior. Average precision falls from $1.0000$ to
$0.8711$, a loss of $12.9$ points against the linear model's $5.2$; prior-matched
\Fone{} falls to $0.6886 \pm 0.0027$ against the linear model's $0.9427$; and
balanced accuracy falls to $0.6131$ against $0.9475$. At its actual operating
point the transformer collapses to \Fone{}~$=0.2592$ with precision $0.1489$ at
recall $0.9992$: it labels almost everything fake, and since the shifted test
prior is $0.119$, a precision of $0.1489$ is only marginally better than a
degenerate always-fake rule. Two qualifications keep this honest. The ranking is
not destroyed---an average precision of $0.8711$ remains far above the $0.119$
expected of a random ranker---and the threshold-oracle value of $0.8023$ shows
that a well-chosen cutoff would recover much of the deployed loss. But every
prior-robust comparison places the transformer well below the linear model
off-distribution, so the reversal is not an artefact of calibration alone.

Temporal shift again shows the opposite pattern: the transformer holds at
\Fone{}~$=0.9922$ and average precision $0.9999$, marginally ahead of the linear
model. This reinforces the conclusion of Section~\ref{sec:results-shift} that a
chronological split is not a meaningful stress test for this corpus, for either
model class.

Truncation does not explain the result. Repeating the random protocol with a
512-token limit changes \Fone{} only from $0.9993$ to $0.9994$
(Table~\ref{tab:transformer}), so the 256-token window is not the binding
constraint on what the model can see.

The most practically alarming number in the table is not a test score but a
validation score. Under the topic-disjoint protocol the selected checkpoint had
validation \Fone{}~$=0.9995$, higher than under the random protocol, while its
test \Fone{} was $0.2592$. Because the validation split must be drawn from the
training pool when no target-domain labels exist, standard model selection
returned a confident, well-validated model that fails almost completely on the
target distribution. No amount of care with the validation set detects this;
only a shifted evaluation does.

\emph{Robustness across seeds.} Table~\ref{tab:transformer} reports a single
representative run (seed 42) for a like-for-like comparison with the
deterministic linear models. To confirm that its topic-shift collapse is not a
lucky draw, we repeat both protocols across five seeds
(Table~\ref{tab:multiseed}). In-distribution performance is essentially constant
(average precision $1.0000 \pm 0.0000$). The topic-disjoint result is stable in
direction and moderate in spread: average precision is $0.8610 \pm 0.0567$ and
prior-matched \Fone{} is $0.6858 \pm 0.0251$, and every one of the five seeds
lands below the linear model's average precision of $0.9475$---the best seed
reaches only $0.9232$. The model-selection trap is the most stable feature of
all: validation \Fone{} is $0.9999 \pm 0.0001$ across seeds while deployed test
\Fone{} averages $0.2575 \pm 0.0235$. The single-seed conclusion therefore holds
under reseeding.

\begin{table}[t]
\centering
\caption{Five-seed distribution (seeds $\{42,0,1,2,3\}$) of the DistilBERT
control, as mean $\pm$ standard deviation. In-distribution scores are constant;
the topic-disjoint collapse is robust, and its validation \Fone{} stays near $1$
in every seed even as test \Fone{} averages $0.26$.}
\label{tab:multiseed}
\footnotesize
\setlength{\tabcolsep}{3.4pt}
\begin{tabular}{@{}lcc@{}}
\toprule
Metric & Random & Topic-disjoint\\
\midrule
Validation \Fone{}          & $0.9991 \pm .0004$ & $0.9999 \pm .0001$\\
Test \Fone{} (deployed)     & $0.9992 \pm .0001$ & $0.2575 \pm .0235$\\
Threshold-oracle \Fone{}    & $0.9994 \pm .0002$ & $0.8049 \pm .0383$\\
Average precision           & $1.0000 \pm .0000$ & $0.8610 \pm .0567$\\
Balanced accuracy           & $0.9993 \pm .0001$ & $0.6059 \pm .0431$\\
Prior-matched \Fone{}       & $0.9992 \pm .0001$ & $0.6858 \pm .0251$\\
ROC-AUC                     & $1.0000 \pm .0000$ & $0.9457 \pm .0332$\\
\bottomrule
\end{tabular}
\end{table}

Taken together, Q4 answers in the direction that the data, not the instrument,
is the binding constraint. Capacity sufficient to represent veracity-bearing
content did not cause the model to use it. Instead the transformer fit the
provenance signature more sharply, reaching an essentially perfect
in-distribution ranking, and paid for it with a steeper loss when that signature
stopped being predictive.

\begin{table*}[t]
\centering
\caption{Capacity control: fine-tuned DistilBERT against the selected linear
model on identical documents and splits. Columns follow
Table~\ref{tab:shift}. The transformer wins in-distribution on every metric and
loses off-distribution on every prior-robust metric. The final row repeats the
random protocol at a 512-token limit, confirming that truncation is not the
operative factor. Validation \Fone{} is shown to expose the model-selection
trap: under topic shift it is at its highest exactly where test performance is
at its worst. DistilBERT rows are the seed-42 run; the five-seed distribution is
in Table~\ref{tab:multiseed}.}
\label{tab:transformer}
\footnotesize
\begin{tabular}{@{}llccccccccc@{}}
\toprule
Protocol & Model & Val.\ \Fone{} & $P$ & $R$ &
$F_1^{\text{def}}$ & $F_1^{\text{orc}}$ & AUC & AP & Bal.\ acc. &
$F_1^{\text{match}}$\\
\midrule
\multirow{2}{*}{Random (standard)}
 & TF--IDF $+$ PA & 0.9924 & 0.9933 & 0.9877 & 0.9905 & 0.9918 & 0.9995 & 0.9995 & 0.9910 & $0.9905$\\
 & DistilBERT     & 0.9992 & \textbf{0.9997} & \textbf{0.9989} & \textbf{0.9993} & \textbf{0.9997} & \textbf{1.0000} & \textbf{1.0000} & \textbf{0.9993} & $\mathbf{0.9993}$\\
\midrule
\multirow{2}{*}{Topic-disjoint}
 & TF--IDF $+$ PA & 0.9928 & \textbf{0.7011} & 0.9498 & \textbf{0.8067} & \textbf{0.8887} & \textbf{0.9872} & \textbf{0.9475} & \textbf{0.9475} & $\mathbf{0.9427}${\scriptsize$\pm.0032$}\\
 & DistilBERT     & 0.9995 & 0.1489 & \textbf{0.9992} & 0.2592 & 0.8023 & 0.9654 & 0.8711 & 0.6131 & $0.6886${\scriptsize$\pm.0027$}\\
\midrule
\multirow{2}{*}{Temporal}
 & TF--IDF $+$ PA & 0.9684 & \textbf{0.9670} & 0.9880 & 0.9774 & 0.9809 & 0.9993 & 0.9964 & 0.9918 & $0.9914${\scriptsize$\pm.0013$}\\
 & DistilBERT     & 0.9956 & 0.9857 & \textbf{0.9988} & \textbf{0.9922} & \textbf{0.9946} & \textbf{1.0000} & \textbf{0.9999} & \textbf{0.9985} & $\mathbf{0.9985}$\\
\midrule
Random, 512 tokens & DistilBERT & 0.9997 & 0.9994 & 0.9994 & 0.9994 & 0.9996 & 1.0000 & 1.0000 & 0.9995 & $0.9994$\\
\bottomrule
\end{tabular}
\end{table*}

\subsection{Q5: Cross-Corpus Transfer Collapses to Chance}
\label{sec:results-crosscorpus}
\begin{figure}[t]
\centering
\includegraphics[width=\columnwidth]{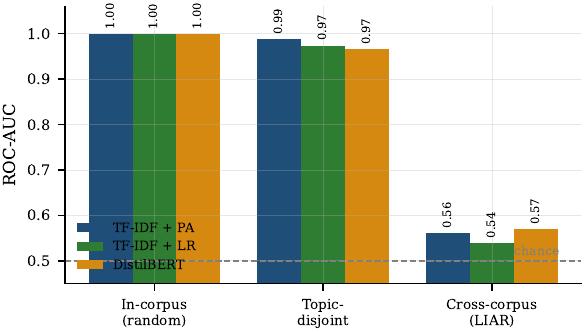}
\caption{ROC-AUC of each model across three evaluation regimes: in-corpus random
split, in-corpus topic-disjoint split, and cross-corpus transfer to LIAR.
Ranking quality survives topic shift within ISOT but collapses to near-chance on
an independently constructed corpus. The dashed line marks chance ($0.5$).}
\label{fig:crosscorpus}
\end{figure}

The within-corpus protocols hold the corpus fixed. The decisive test of external
validity is to change the corpus entirely. We train each model on all
$38{,}826$ ISOT documents and evaluate on the LIAR test split, whose short
single-source claims share almost none of ISOT's provenance structure.
Table~\ref{tab:crosscorpus} reports the result, and Figure~\ref{fig:crosscorpus}
places it beside the in-corpus and topic-shift figures.

The transfer is essentially complete failure. ROC-AUC falls to $0.5398$ for
logistic regression, $0.5617$ for passive--aggressive, and $0.5691$ for
DistilBERT, all within seven points of the chance value of $0.5$; balanced
accuracy lies between $0.5244$ and $0.5487$ against a chance value of $0.5$; and
average precision ($0.4859$ to $0.5060$) barely exceeds the $0.436$ expected of a
random ranker. None of the three models reaches the accuracy of a majority-class
predictor ($0.5635$): trained to near-perfection on ISOT, they are worse than a
constant ``always real'' rule on LIAR. The ordering among them is preserved but
meaningless at this scale---DistilBERT ranks marginally best, as it did
in-distribution, but an AUC of $0.5691$ is not a usable classifier.

Two controls rule out trivial explanations. Every one of the $1{,}267$ LIAR
documents contains at least one term in the ISOT vocabulary, so the collapse is
not an artefact of empty feature vectors; the shared words simply carry no
veracity signal the ISOT model can use. And the effect holds for the transformer
as well as the linear models, so it is not a limitation of bag-of-words
representation. The contrast with the within-corpus AUC values, which remain
near $1.0$ even under topic shift (Table~\ref{tab:shift}), is the sharpest
statement of the paper's thesis: what the models learn ranks ISOT documents
almost perfectly and ranks LIAR documents no better than a coin.

This is the external check that Q1--Q4 could only approach from inside the
corpus. A model can attain \Fone{}~$=0.99$ on this benchmark and carry
essentially no transferable capacity to judge whether a claim is true.

\begin{table}[t]
\centering
\caption{Cross-corpus transfer: trained on all of ISOT, tested on the LIAR test
split ($n=1{,}267$, fake prior $0.436$). All models fall to near-chance ranking
(AUC and balanced accuracy near $0.5$; average precision near the prior $0.436$)
and none beats the majority-class accuracy of $0.5635$. Compare the
within-corpus AUC values near $1.0$ in Table~\ref{tab:shift}.}
\label{tab:crosscorpus}
\footnotesize
\setlength{\tabcolsep}{3.4pt}
\begin{tabular}{@{}lcccccc@{}}
\toprule
Model & Acc. & $P$ & $R$ & \Fone{} & AUC & Bal.\ acc.\\
\midrule
Majority class (real) & 0.5635 & -- & 0.000 & 0.000 & -- & 0.5000\\
\midrule
TF--IDF $+$ logistic reg. & 0.5067 & 0.4553 & 0.6637 & 0.5401 & 0.5398 & 0.5244\\
TF--IDF $+$ pass.-aggr.   & 0.5201 & 0.4670 & 0.7034 & 0.5613 & 0.5617 & 0.5408\\
DistilBERT                & 0.5375 & 0.4776 & 0.6365 & 0.5457 & 0.5691 & 0.5487\\
\bottomrule
\end{tabular}
\end{table}

\subsection{Computational Cost}
\label{sec:results-cost}
Table~\ref{tab:cost} reports measured cost on the hardware of
Section~\ref{sec:setup}. Vectorising $27{,}177$ documents takes $12.19$\,s and
dominates training, which needs $0.44$\,s for logistic regression and under
$0.2$\,s for the other three; the full four-model pipeline completes in well
under a minute on a single CPU. The serialised pipeline occupies $2.15$\,MB and
sustains $3{,}002$ documents per second in batch mode, with $0.66$\,ms
single-document latency. Two consequences follow. First, the audit is
inexpensive: the complete linear study, comprising 20 fitted conditions, runs in
roughly twelve minutes on one CPU, and the DistilBERT control adds about
32 minutes on one laptop GPU. The diagnostics we recommend therefore impose no
practical barrier. Second, on cost grounds the linear pipeline would be an
attractive deployment candidate---it needs no accelerator, and it holds the
better off-distribution operating point of the two arms---which is precisely why
its failure under topic shift matters.
Scalability is favourable in the regime that matters here: memory grows as
$O(Nz)$ rather than $O(NV)$, and inference is independent of $N$, so
throughput would be maintained on substantially larger corpora, with
vocabulary construction the first component to require distribution.

\begin{table}[t]
\centering
\caption{Measured computational cost of the linear pipeline (single CPU, no
GPU; $N=27{,}177$ training documents, $V=50{,}000$). The DistilBERT control,
for comparison, requires $5.8$--$6.5$ minutes of GPU time per protocol.}
\label{tab:cost}
\footnotesize
\begin{tabular}{@{}lr@{}}
\toprule
Quantity & Value\\
\midrule
Vectorisation (fit $+$ transform train) & $12.19$\,s\\
Training: logistic regression / MNB & $0.44$\,s / $0.016$\,s\\
Training: LinearSVC / passive--aggressive & $0.16$\,s / $0.091$\,s\\
Non-zeros in training matrix & $5.17\times10^{6}$\\
Mean non-zeros per document ($z$) & $190.1$ ($0.38\%$ density)\\
Sparse / dense memory for training matrix & $\approx62$\,MB / $\approx10.9$\,GB\\
Serialised pipeline size & $2.15$\,MB\\
Batch throughput & $3{,}002$ docs/s\\
Single-document latency & $0.66$\,ms\\
\bottomrule
\end{tabular}
\end{table}

\section{Discussion}
\label{sec:discussion}

\subsection{What the Benchmark Score Measures}
Consider the evidence together. The corpus can be solved exactly from a
metadata field (Section~\ref{sec:results-ablation}), and removing that field
along with two further leakage channels costs only $1.21$ points. What survives
is insensitive to representation, acquirable from a few hundred examples, and
robust to deleting a thousand of its strongest features
(Section~\ref{sec:results-representation}); the features it does rely on are
recognisably editorial conventions (Figure~\ref{fig:coefficients}). Finally, it
degrades measurably when provenance and topic change, but not when only time
changes (Section~\ref{sec:results-shift}). The consistent explanation is that
high scores on this benchmark quantify \emph{source and topic separability}. In
the notation of Section~\ref{sec:problem}, the pipeline is an efficient
estimator of $s$, and because $I(s;y)$ is near-maximal in
$\mathcal{P}_{\text{bench}}$ by the asymmetry of Equation~\eqref{eq:mi}, an
estimator of $s$ is nearly an estimator of $y$ in-distribution. Nothing in our results indicates that veracity-bearing content
$c$ is being used.

This is a claim about the benchmark, not about the feasibility of the task. It
also does not imply prior work is invalid: reported numbers are reproducible
under the protocols used. What does not follow is the inference from those
numbers to deployment readiness.

\subsection{Why Model Capacity Does Not Help}
The capacity control of Section~\ref{sec:results-transformer} lets us make this
argument from our own evidence rather than by inference from the literature's
near-ceiling clustering \cite{khan2021benchmark,kaliyar2021fakebert}. A
pretrained transformer, able in principle to represent the veracity-bearing
content $c$ that a bag of words cannot, did not use it. It fit the provenance
signature $s$ more sharply instead, achieving an essentially perfect
in-distribution ranking, and then lost $2.5$ times more average precision than
the linear model when $s$ ceased to be predictive. Capacity was spent on the
shortcut, not on an alternative to it.

The explanation is that the limitation is an identifiability problem in the data
rather than a modelling deficiency. $\mathcal{P}_{\text{bench}}$ contains no
counterfactual pairs (same publisher, different veracity), so no objective
computed on it can distinguish a rule keyed on $c$ from a rule keyed on $s$.
Both achieve identical empirical risk. Gradient descent then selects whichever is
easier to extract, and provenance style is trivially easier than factual
verification. A higher-capacity model does not resolve this ambiguity; it merely
resolves the easy side of it more precisely, which is why its
in-distribution advantage and its out-of-distribution disadvantage appear
together. This is the mechanism Schuster et al.
\cite{schuster2020limitations} describe for machine-generated text, and our
metadata-only and deletion probes are the analogue of hypothesis-only
diagnostics in NLI \cite{gururangan2018annotation,poliak2018hypothesis}.

There is a corollary worth stating plainly, because it inverts a common
assumption. On a benchmark whose labels are confounded with provenance, a
\emph{higher} reported score is weak evidence of a better system and can be
evidence of a worse one. Our transformer would be selected over the linear model
by any standard random-split comparison, and it is the worse choice for
deployment on unseen sources.

\subsection{Metric Choice and Deployment Risk}
The shift analysis carries a practical lesson beyond this corpus. Under
topic-disjoint evaluation, ROC-AUC remains at $0.9872$ for the selected model
while deployed \Fone{} sits at $0.8067$ (Figure~\ref{fig:roc}). A study
reporting only ROC-AUC would therefore describe a system that is, at its actual
operating point, wrong about $30\%$ of its positive predictions. Because
misinformation is rare in realistic streams, precision as defined in
Equation~\eqref{eq:prf} is the quantity that governs harm, since false positives
suppress legitimate journalism, and it is exactly the quantity most sensitive to
the prior shift that accompanies deployment. We therefore recommend reporting average precision
together with the operating point actually used, and treating calibration as a
first-class concern rather than a post-hoc adjustment
\cite{platt1999probabilistic,niculescu2005predicting,guo2017calibration}.

\subsection{Recommended Diagnostics}
The probes used here are cheap, a few minutes of CPU time each, and we suggest
them as routine reporting for any study using this corpus or one built by
pairing publisher sets:
\begin{enumerate}
  \item \textbf{Metadata-only baseline.} Train on non-textual fields alone. If
    it solves the task, textual results on the same feature set are
    uninterpretable.
  \item \textbf{Small-sample baseline.} Report performance at $1$--$2\%$ of the
    training data. Near-ceiling performance there indicates a shallow signal.
  \item \textbf{Duplicate audit.} Report the fraction of test documents
    occurring verbatim in training, before de-duplication.
  \item \textbf{Topic- or source-disjoint split.} Report alongside the random
    split. Our results indicate this is a far more sensitive stress test than a
    temporal split for this corpus.
  \item \textbf{Prior-controlled reporting.} When protocols change the class
    prior, report average precision and prior-matched \Fone{} so that
    calibration effects are not mistaken for lost discrimination.
  \item \textbf{Do not trust in-domain validation as a shift warning.} Our
    transformer's validation \Fone{} was $0.9995$---its highest across
    protocols---in the very condition where test \Fone{} was $0.2592$. A
    validation split drawn from the training pool, which is all that is
    available without target labels, provides no signal about this failure.
    Report a shifted evaluation or report nothing about generalisation.
\end{enumerate}

\section{Limitations}
\label{sec:limitations}

We state the boundaries of these conclusions explicitly.

\textbf{Training corpus and label mapping.} All models are trained on one
corpus, and our conclusions concern ISOT and, by extension, corpora built by
pairing disjoint publisher sets. The cross-corpus test
(Section~\ref{sec:results-crosscorpus}) uses a single external benchmark, LIAR,
with one binarisation of its six-way labels; a different cut, or a different
external corpus such as NELA-GT \cite{norregaard2019nela}, could shift the exact
numbers, though the near-chance transfer is a wide margin that a boundary choice
is unlikely to close. LIAR's claim-level format also differs from ISOT's
article-level format, so the transfer failure conflates a change of source with
a change of granularity; disentangling the two would require an article-level
external corpus.

\textbf{Instrument capacity.} The primary probe is a bag-of-words linear model
that cannot represent word order beyond bigrams, negation, or discourse
structure. The DistilBERT control (Section~\ref{sec:results-transformer})
addresses the obvious objection, but it does not license an unlimited claim. We
show that one widely used pretrained model, under a standard fine-tuning recipe,
exploits the shortcut more rather than less; we do not show that no model or
training procedure could learn veracity-bearing features from this corpus. Our
claim remains asymmetric in that direction.

\textbf{Scope of the capacity control.} The transformer result rests on a single
architecture, two epochs, and no hyperparameter search, though it is averaged
over five seeds (Table~\ref{tab:multiseed}). The topic-disjoint average precision
carries real spread ($0.8610 \pm 0.0567$), so individual magnitudes should be
read with that variance in mind; the direction, however, is consistent across
all five seeds and across three prior-robust metrics, and every seed falls below
the linear model. A larger model and a search over learning rate and epoch count
would sharpen the estimate further, and a single architecture cannot speak for
all transformers.

\textbf{Asymmetric preprocessing in the control.} The transformer receives
minimally normalised text while the linear model receives alpha-only text
(Section~\ref{sec:method-transformer}). Document sets, splits, and leakage
controls are identical, but this surface difference means the two arms are not
preprocessed identically. It favours the transformer, so it cannot explain the
transformer's worse off-distribution result; it does inflate its
in-distribution advantage by an amount we have not measured.

\textbf{GPU nondeterminism.} Unlike the linear pipeline, the transformer runs
mixed-precision CUDA kernels and is therefore reproducible only to
approximately three decimal places.

\textbf{Confounded shift protocol.} Subject values proxy simultaneously for
topic, sub-publication, and time-of-collection, so the topic-disjoint protocol
perturbs several factors at once. Our prior-matched analysis removes the
class-prior confound but not this one; the resulting AP drop should be read as
the effect of a bundle of provenance-related shifts rather than of topic alone.
A source-level split with publisher metadata---absent from this
distribution---would be cleaner.

\textbf{Exact-duplicate detection only.} De-duplication matches identical
cleaned strings. Near-duplicates (syndicated rewrites, partial overlaps) are
not detected, so the $19.37\%$ contamination figure is a lower bound and the
$0.9814$ primary score may still be mildly optimistic.

\textbf{Residual artifacts.} We remove one agency name. Other provenance traces
(wire-service phrasing, content-management boilerplate) remain and are, by the
argument of Section~\ref{sec:results-representation}, precisely what the model
uses. We do not claim to have produced an artifact-free version of this corpus;
we claim the opposite, namely that doing so by token filtering is not feasible.

\textbf{English, single period, no external evidence.} The corpus is English
and concentrated in 2016--2017 US politics, and the pipeline consults no
knowledge source, so it cannot in principle verify a claim.

\textbf{Statistical scope.} Bootstrap intervals and McNemar tests quantify
uncertainty on a fixed test split; cross-validation quantifies split
sensitivity. Neither accounts for variation across corpus construction choices,
which our ablations address descriptively rather than inferentially.

\section{Future Work}
\label{sec:futurework}

\textbf{Article-level cross-corpus transfer.} Our LIAR result
(Section~\ref{sec:results-crosscorpus}) changes both source and granularity at
once, since LIAR is claim-level. Repeating the transfer test on an article-level
external corpus such as NELA-GT \cite{norregaard2019nela} would isolate the
effect of changing source alone, and would test whether the near-chance collapse
we observe is specific to LIAR's short claims or general to out-of-corpus text.

\textbf{Source-controlled data construction.} The identifiability problem
motivates corpora containing veracity variation \emph{within} publisher, so
that $s$ is uninformative about $y$ by design. Retracted or corrected articles
from a single outlet are one route; matched-pair sampling on topic and outlet is
another.

\textbf{Scaling the capacity control.} Our DistilBERT result
(Section~\ref{sec:results-transformer}) is a single architecture at a single
seed. Extending it to a multi-seed protocol, to full-size encoders, and to
instruction-tuned decoder models would establish whether the amplification we
observe grows or diminishes with scale---the question that matters for whether
current large models are safe to trust on provenance-confounded benchmarks.

\textbf{Debiasing and invariance.} Adversarial removal of source-predictive
directions, reweighting, or invariant-risk objectives could be evaluated by
whether they improve topic-disjoint AP without degrading in-distribution
performance---the trade-off our protocol is designed to expose.

\textbf{Calibration under shift.} Given that most of the deployed \Fone{}
collapse was calibration, methods that maintain calibration without target
labels \cite{guo2017calibration} deserve evaluation in this setting.

\textbf{Beyond exact duplicates.} Near-duplicate detection (shingling, MinHash)
would tighten the contamination estimate and yield a cleaner corpus version.

\section{Conclusion}
\label{sec:conclusion}

We audited a widely used fake news corpus by holding a transparent TF--IDF and
linear-classifier pipeline fixed and varying the data and the evaluation
protocol. The pipeline reproduces the literature's near-ceiling result
(\Fone{}~$=0.9905$, 95\% CI $[0.9883,0.9928]$), and we then established what
that number rests on. A metadata-only classifier that never reads the article
text solves the benchmark exactly (\Fone{}~$=1.000$). Removing that field, a
newswire source tag present in $99.21\%$ of real articles, and duplicates that
contaminate $19.37\%$ of a naive test split costs only $1.21$ \Fone{} points,
because the residual signal is diffuse editorial style: nine representation
variants span $0.62$ points, $543$ documents recover $95.1\%$ of full-data
performance, and deleting the $1{,}000$ highest-weight unigrams leaves \Fone{}
at $0.9263$. That signal does not transfer. Under a topic-disjoint protocol
average precision falls from $0.9995$ to $0.9475$ and deployed \Fone{} to
$0.8067$; a prior-matched analysis attributes most of the latter to class-prior
shift and miscalibration ($0.9427$) while confirming a genuine $5.2$-point loss
of discrimination. Temporal transfer is by contrast nearly lossless, which
identifies topic- and source-disjoint splits---not chronological ones---as the
informative stress test for this corpus.

A capacity control settles where the limitation lies. Fine-tuned DistilBERT,
trained on identical documents and splits, is the stronger classifier
in-distribution (\Fone{}~$=0.9993$, average precision $1.0000$) and the weaker
one under topic shift on every prior-robust metric, losing $12.9$ points of
average precision against the linear model's $5.2$ and reaching prior-matched
\Fone{} of $0.6886$ against $0.9427$. Capacity sufficient to represent
veracity-bearing content was instead spent fitting the provenance signature more
sharply. Notably, its validation \Fone{} peaked at $0.9995$ in precisely the
condition where test \Fone{} fell to $0.2592$, so standard in-domain model
selection gives no warning of the failure. The cross-corpus test is the final
word: transferred to the LIAR benchmark, all three ISOT-trained models rank at
near-chance (ROC-AUC $0.54$--$0.57$) and none beats a majority-class predictor,
even though every model exceeds $0.99$ AUC within ISOT. A benchmark score of
$0.99$ on this corpus therefore certifies almost no transferable ability to
judge veracity.

Our contribution is measurement rather than modelling, and its limits are
correspondingly specific: one corpus, a shift protocol that perturbs several
provenance factors together, and a capacity control at one architecture and one
seed (Section~\ref{sec:limitations}). Within those limits the practical
implication is concrete. Reported accuracy on this benchmark should be read as a measurement
of source and topic separability, and studies using it should accompany their
headline number with the metadata-only, small-sample, duplicate-audit, and
topic-disjoint diagnostics of Section~\ref{sec:discussion}, each of which costs
minutes of CPU time. For practitioners, the operative finding is that a system
scoring $0.99$ in-distribution can mispredict roughly $30\%$ of its positive
decisions once the topic mix changes---a failure mode invisible to
threshold-free metrics and to random-split evaluation alike. All code, seeds,
and derived results are released to make each of these checks a matter of
re-execution rather than reimplementation.

\appendices

\section{Hyperparameters}
\label{app:hyper}
Table~\ref{tab:hyper} lists every setting required to reproduce the reported
numbers. Values are not tuned: defaults were fixed before the experiments to
avoid an implicit search that the single fixed test split could not support. A
consequence is that reported figures are not upper bounds; the ablations of
Table~\ref{tab:representation} show the achievable range is narrow in any case.

\begin{table}[h]
\centering
\caption{Complete hyperparameter listing.}
\label{tab:hyper}
\footnotesize
\begin{tabular}{@{}llp{2.85cm}@{}}
\toprule
Component & Parameter & Value\\
\midrule
\multirow{7}{*}{TF--IDF}
 & \texttt{ngram\_range}  & $(1,2)$\\
 & \texttt{min\_df}       & 5\\
 & \texttt{max\_df}       & 0.9\\
 & \texttt{max\_features} & 50{,}000\\
 & \texttt{sublinear\_tf} & true\\
 & \texttt{stop\_words}   & English (built-in list)\\
 & \texttt{strip\_accents}& unicode\\
\midrule
Logistic regression & $C$, solver, \texttt{max\_iter} & $1.0$, liblinear, 1000\\
Multinomial NB      & $\alpha$ & $0.1$\\
LinearSVC           & $C$, loss & $1.0$, squared hinge\\
Passive--aggressive & $C$, \texttt{max\_iter} & $1.0$, 1000\\
\midrule
\multirow{8}{*}{DistilBERT}
 & checkpoint & \texttt{distilbert-}\\
 &            & \texttt{base-uncased}\\
 & max sequence length & 256 (512 in the\\
 &                     & sensitivity run)\\
 & batch size, epochs & 16, 2\\
 & learning rate, warmup & $2\times10^{-5}$, 10\% linear\\
 & precision & fp16 autocast\\
 & selection & best per-epoch validation \Fone{}\\
\midrule
\multirow{4}{*}{Protocol}
 & split & $70/10/20$, stratified\\
 & seed $\sigma$ & 42\\
 & min.\ document length & 20 characters\\
 & selection metric & validation \Fone{}\\
\midrule
\multirow{2}{*}{Statistics}
 & bootstrap resamples & 2000\\
 & prior-match repeats $M$ & 20\\
\bottomrule
\end{tabular}
\end{table}

\section{Learning-Curve and Deletion-Probe Values}
\label{app:lc}
Numerical values underlying Figure~\ref{fig:learning} (logistic-regression
probe, primary configuration, random protocol), as training documents
$\rightarrow$ test \Fone{}:
$135 \rightarrow 0.6802$;
$271 \rightarrow 0.8704$;
$543 \rightarrow 0.9334$;
$1{,}358 \rightarrow 0.9370$;
$2{,}717 \rightarrow 0.9579$;
$6{,}794 \rightarrow 0.9714$;
$13{,}588 \rightarrow 0.9765$;
$27{,}177 \rightarrow 0.9814$.

Values underlying Figure~\ref{fig:deletion}, as unigrams removed $K$
$\rightarrow$ test \Fone{}:
$0 \rightarrow 0.9814$;
$10 \rightarrow 0.9758$;
$50 \rightarrow 0.9662$;
$100 \rightarrow 0.9600$;
$500 \rightarrow 0.9383$;
$1000 \rightarrow 0.9263$.
The twenty highest-weight unigrams, removed first, are: \emph{said, video,
washington, image, just, featured, read, gop, wednesday, watch, tuesday, obama,
america, thursday, hillary, getty, like, friday, republican, minister}.

\section{Corpus Composition}
\label{app:corpus}
\begin{figure}[h]
\centering
\includegraphics[width=\columnwidth]{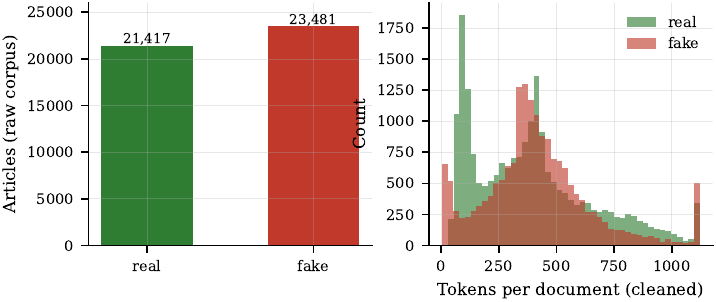}
\caption{Raw class counts (left) and token-length distributions after
normalisation, clipped at the 98th percentile for legibility (right).}
\label{fig:corpus}
\end{figure}

Figure~\ref{fig:corpus} shows class counts in the raw distribution and the
token-length distribution after normalisation. The length distributions overlap
substantially, so length alone is a weak cue; the bimodality visible in the real
class reflects the mixture of short wire briefs and full reports.

\section{Reproducibility Checklist}
\label{app:checklist}
\begin{itemize}
  \item \textbf{Code.} Released at
    \url{https://github.com/vermayuvraj/fake-news-detection}, including the
    audit scripts that regenerate every table and figure.
  \item \textbf{Data.} Not redistributed; obtained from
    \cite{bisaillon2020dataset}. Construction from the raw files is fully
    specified by Algorithm~\ref{alg:corpus}.
  \item \textbf{Environment.} Exact versions pinned
    (Section~\ref{sec:setup}); no GPU required.
  \item \textbf{Randomness.} Single seed $\sigma=42$ for splits and models;
    deterministic solver and frequency-ordered vocabulary. Two independent runs
    of the linear pipeline produced bit-identical metrics; the DistilBERT
    control is seeded identically but, running mixed-precision CUDA kernels,
    reproduces to approximately three decimal places.
  \item \textbf{Split alignment.} The transformer script asserts that its split
    cardinalities equal those of the linear runs before training, so the
    capacity comparison cannot silently drift out of alignment.
  \item \textbf{Protocol integrity.} Model selection on validation only; the
    test split is evaluated once per condition; the vectoriser is fitted on
    training data only.
  \item \textbf{Negative control.} Label-permutation run yields test
    ROC-AUC $=0.4974$.
  \item \textbf{Uncertainty.} Bootstrap 95\% intervals ($B{=}2000$), five-fold
    cross-validation, McNemar exact tests, and $M{=}20$ prior-matched
    resamples.
  \item \textbf{Artifacts.} Machine-readable results
    (\texttt{results.json}, \texttt{shift\_analysis.json},
    \texttt{transformer.json}, \texttt{transformer\_multiseed.json},
    \texttt{crosscorpus.json}) accompany the code, and all reported values are
    drawn from them.
  \item \textbf{Compute.} Linear study $\approx12$ minutes on one CPU;
    transformer control $\approx32$ minutes on one RTX~4050 (three protocols
    at $6.5$, $5.8$ and $6.2$ minutes, plus $13.0$ minutes for the 512-token
    replication); five-seed protocol a further $\approx60$ minutes; cross-corpus
    evaluation under one minute plus the LIAR download.
\end{itemize}

\section{Use of AI Tools}
\label{app:aitools}
Consistent with arXiv policy, we disclose tool use. Modelling, data analysis,
and figures use the standard scientific-Python stack cited in
Section~\ref{sec:setup}; no generative model is part of the method, the data
pipeline, or the results. During manuscript preparation, an AI coding assistant
was used to help implement the experiment scripts and to help draft and edit
prose. All experiments were executed by the author on the author's hardware;
every reported number is produced by the released code and independently checked
against the machine-readable result files by a verification script also included
in the repository. The author reviewed all code and text and takes full
responsibility for the contents of this paper, including any errors.

\bibliographystyle{IEEEtran}
\bibliography{refs}

\end{document}